\documentclass[runningheads]{llncs}

\usepackage{eccv}

\usepackage{eccvabbrv}

\usepackage{graphicx}
\usepackage{booktabs}
\usepackage{subcaption}

\usepackage[dvipsnames]{xcolor}
\usepackage{algorithm}
\usepackage{algpseudocode}
\usepackage{multirow}
\usepackage{titletoc}
\usepackage{wrapfig}
\usepackage{subfloat}
\usepackage{graphicx}
\usepackage{tabularx}
\usepackage{pifont}
\usepackage{array}
\usepackage{tocloft}
\usepackage{amsmath} 
\DeclareMathOperator*{\argmin}{arg\,min} 
\DeclareMathOperator*{\argmax}{arg\,max} 
\newcolumntype{C}[1]{>{\centering\arraybackslash}p{#1}}

\newcommand{\xmark}{\ding{55}}

\usepackage[accsupp]{axessibility}  

\usepackage{hyperref}

\usepackage{orcidlink}

\hypersetup{
  colorlinks   = true, 
  citecolor   = black 
}

\begin{document}

\title{Understanding Physical Dynamics with Counterfactual World Modeling} 

\titlerunning{Counterfactual World Modeling}

\author{
	Rahul Venkatesh\inst{1}\textsuperscript{$*$}\and
        Honglin Chen\inst{1}\textsuperscript{$*$}\and
        Kevin Feigelis\inst{1}\textsuperscript{$*$}\and
        Daniel M. Bear\inst{1}\and
        Khaled Jedoui\inst{1} \and
        Klemen Kotar\inst{1} \and
        Felix Binder\inst{3} \and
        Wanhee Lee\inst{1} \and
        Sherry Liu\inst{1}\ \and
        Kevin A. Smith\inst{2} \and
        Judith E. Fan\inst{1} \and
        Daniel L. K. Yamins\inst{1} 
}
\authorrunning{Venkatesh et al.}
\institute{        
\textsuperscript{$1$} Stanford \quad \quad
\textsuperscript{$2$} MIT \quad \quad
\textsuperscript{$3$} UC San Diego}
\def\thefootnote{*}\footnotetext{Equal contribution}\def\thefootnote{\arabic{footnote}}
\maketitle

\begin{abstract}


The ability to understand physical dynamics is critical for agents to act in the world. Here, we use Counterfactual World Modeling (CWM) to extract vision structures for dynamics understanding. CWM uses a temporally-factored masking policy for masked prediction of video data without annotations. This policy enables highly effective ``counterfactual prompting'' of the predictor, allowing a spectrum of visual structures to be extracted from a single pre-trained predictor without finetuning on annotated datasets. We demonstrate that these structures are useful for physical dynamics understanding, allowing CWM to achieve the state-of-the-art performance on the Physion benchmark. Project Website: \href{https://neuroailab.github.io/cwm-physics/}{\texttt{\textbf{https://neuroailab.github.io/cwm-physics/}}}.

\end{abstract}

\section{Introduction}
\label{sec:intro}

Physical dynamics understanding involves predicting the effects of physical interactions with objects (e.g. predicting the trajectory of a thrown ball \cite{gerstenberg2017eye}, or the direction of a falling stacked block tower \cite{battaglia2013simulation}).  This remains a critical challenge for autonomous agents such as robots and self-driving cars interacting with the world \cite{finn2016unsupervised}. Existing computer vision algorithms significantly lag behind humans in physical dynamics understanding \cite{bear2021physion}.


One class of existing methods relies on intermediate vision structures such as 2D object segmentations and 3D particle graphs \cite{battaglia2016interaction,mottaghi2016newtonian, li2018learning,bates2019modeling,tacchetti2018relational,sanchez2018graph,mrowca2018flexible,ajay2019combining,smith2019modeling,ye2019compositional,qi2020learning}. These vision structures are highly useful for accurate dynamics prediction because they abstract away irrelevant details. However, these ground-truth structures are only available in simulated or manually annotated datasets. Scaling these approaches to unlabelled real-world video data remains challenging. 




A contrasting class of approaches avoids the use of intermediate structures by learning to predict raw pixels of future video frames \cite{agrawal2016learning,finn2017deep,babaeizadeh2021fitvid,hafner2019learning,fragkiadaki2015learning,hafner2019dream,piloto2022intuitive,wu2022slotformer}. While these approaches are directly applicable to real-world videos, learning to predict future frame pixels poses many challenges due to the high-dimensionality of image pixels and the stochasticity of real-world physical dynamics. These unstructured methods substantially underperform approaches with direct access to ground-truth intermediates, especially 3D particles \cite{bear2021physion}.


Beyond task-specific methods for physical dynamics prediction, a promising alternative is self-supervised learning of task-agnostic visual representations that transfer well to downstream vision tasks \cite{caron2021emerging,he2022masked,tong2022videomae,feichtenhofer2022masked,oquab2023dinov2}. Methods such as DINO \cite{caron2021emerging, oquab2023dinov2}, masked autoencoder (MAE) \cite{he2022masked}, and VideoMAE \cite{feichtenhofer2022masked,tong2022videomae,wang2023videomae} could potentially learn representations useful for dynamics understanding. An additional promise is the emergence of semantic segmentation structure in DINO\cite{caron2021emerging, oquab2023dinov2}, which could potentially improve dynamics understanding. However, these models are mostly used in a transfer learning or fine-tuning paradigm, which requires annotations. It remains unclear whether they can be prompted to extract meaningful structures without finetuning on annotated datasets.

\begin{figure*}[t]
    \centering
    \includegraphics[width=0.9\linewidth]{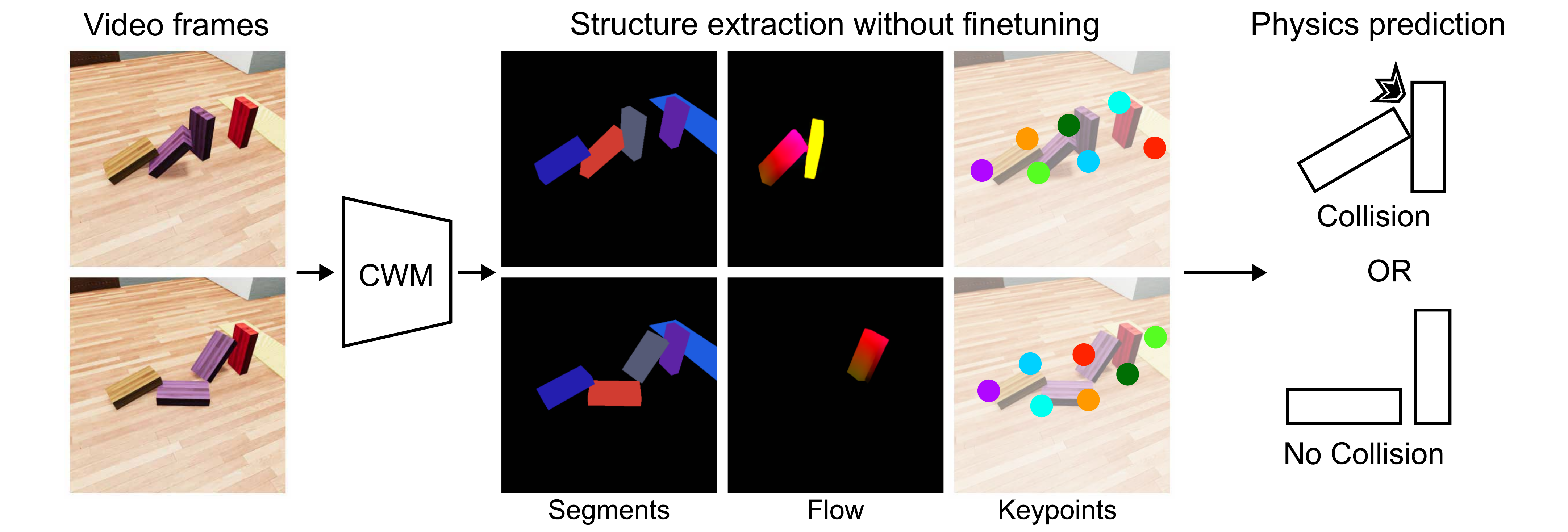} 
    \caption{\textbf{Overview of the approach.} Given an input video of a physical scenario, we extract feature representations and vision structures such as keypoints, optical flow, and segments. These structures are extracted from a single pre-trained CWM predictor without finetuning on annotated datasets. We use the extracted features and structures for dynamics understanding - detecting a past collision or predicting a future collision.}
    \label{fig:overview}
\end{figure*}
Therefore, a key research question is designing methods that pre-train on real-world video data without annotations and support extraction of structures for dynamics understanding. In this work, we use a simple and powerful framework, called Counterfactual World Modeling (CWM)\cite{bear2023unifying}. CWM allows 
extraction of structures useful for understanding dynamics. Figure \ref{fig:overview} provides an overview of our approach. We summarize the contributions of CWM below:

(a) We show that using a \emph{temporally-factored masking policy} during pre-training enables powerful prompting abilities. As in VideoMAE, we train a masked predictor on real-world video data. Unlike VideoMAE, in CWM, the predictor only takes in a few patches of the last frame and fully visible preceding frames as inputs, and predicts the remaining patches in the last frame.  This temporally-factored masking policy encourages the predictor to concentrate information about transformations between frame pairs into the embeddings of a small number of patch tokens. This in turn enables the predictor to support effective prompting via simple interventions on those few key tokens, allowing the system to answer hypotheticals, such as what will the next frame look like if an object in an image is moved to the right. 


(b) We demonstrate that CWM can be prompted to extract multiple vision structures useful for understanding dynamics. As a result of the masking policy, we can extract structures by feeding CWM different prompts. These structures are extracted from a single predictor without being supervised on annotated datasets. Utilizing the extracted structures, CWM achieves state-of-the-art performance on the challenging Physion benchmark \cite{bear2021physion}.

CWM can be understood in the context of Pearl's Ladder of Causation \cite{goldberg2019book}, describing how counterfactual reasoning can be built up from statistical models. The first rung of the Ladder is \emph{Association}, in which a model of the predictable statistical relations between observed events over time is constructed. In CWM, this role is played by the world model itself, the large pretrained predictor which absorbs correlations from observed video inputs. The second rung is \emph{Intervention}, in which at key junctions of the statistical model, observational data are replaced by specific fixed choices (``interventions'') intended to produce some desired outcome. In CWM, this role is played by patch-level prompting, whose utility is greatly enabled by the temporally-factored training of the underlying predictor. The third rung of the Ladder is \emph{Counterfactual}, in which the results of interventions are compared to alternative futures to identify true causes of events. In CWM, the comparison between outcomes of counterfactual interventions (prompts) and alternative futures (observed ground truth or observed predictions) are used for structure extraction, which -- since they better capture core underlying causes of physical events -- end up being useful for improved physical prediction. 

In what follows, we review the literature on related works, and describe the core concepts of the CWM framework. We then demonstrate that the extracted structures of CWM are highly useful for physical dynamics understanding. Lastly, we provide an analysis of the quality of the extracted structures and ablation studies of CWM.

\section{Related Works}
\label{sec:related_works}

\subsubsection{Structured dynamics prediction}
Researchers have made substantial progress in physical dynamics prediction using structured particle representations as inputs \cite{battaglia2016interaction,mottaghi2016newtonian, li2018learning,bates2019modeling,tacchetti2018relational,sanchez2018graph,mrowca2018flexible,ajay2019combining,smith2019modeling, han2022learning}. These approaches simulate large systems of particle-based representations by constructing interaction graphs and propagating information between graph nodes. Besides particle graphs, alternative object structures such as entity locations~\cite{ye2019compositional} and keypoints~\cite{janny2022filtered} are useful for physical dynamics prediction. However, these methods rely on ground-truth object structures, which are only available in simulated or manually annotated datasets. The scalability of these methods on real-world unlabelled data remains limited.
\vspace{-0.5cm}

\subsubsection{Video prediction}
One class of approaches learns physics understanding by predicting the pixels of future video frames \cite{agrawal2016learning,finn2017deep,babaeizadeh2021fitvid,hafner2019learning,fragkiadaki2015learning,hafner2019dream,piloto2022intuitive,wu2022slotformer}. These methods are directly applicable to real-world videos without depending on ground-truth object structures, which are difficult to obtain in general scenarios. Recent video diffusion models \cite{voleti2022mcvd,lu2023vdt, hoppe2022diffusion, chen2023seine} and transformer-based prediction models \cite{gupta2022maskvit, yu2023magvit} have made progress towards more realistic pixel prediction of future video frames. However, learning to predict future frame pixels poses many challenges due to the high-dimensionality of image pixels and the stochasticity of real-world physical dynamics. Existing state-of-the-art methods are prone to creating physically implausible motions in the predicted video frames \cite{liu2024sora}.

\vspace{-0.5cm}

\subsubsection{Self-supervised visual representation learning} Beyond task-specific methods for physical dynamics prediction, a promising alternative is self-supervised learning of task-agnostic visual representations from large-scale unlabeled image or video data. These methods learn to generate visual features that transfer well to downstream vision tasks. One school of works leverages different pretext tasks for pre-training \cite{doersch2015unsupervised,noroozi2016unsupervised,wang2015unsupervised,zhang2016colorful,pathak2017learning,gidaris2018unsupervised}. Another class of works models image similarity and dissimilarity between augmented
views of an image \cite{wu2018unsupervised,oord2018representation,he2020momentum,chen2020simple,oquab2023dinov2, caron2021emerging} and different clips of a video \cite{qian2021spatiotemporal,zhuang2020unsupervised,dave2022tclr} via constrastive learning. The most recent family of masked visual modeling approaches learns effective visual representations via masking and reconstruction of visual tokens. iGPT \cite{chen2020generative} and ViT \cite{dosovitskiy2020image} pioneer this direction by training transformers on pixel or patch tokens and exploring masked prediction with patches. MAE \cite{he2022masked} introduces autoencoding with an asymmetric encoder-decoder architecture and empirically shows that a high masking ratio is crucial for image tasks. VideoMAE \cite{tong2022videomae,feichtenhofer2022masked} extends to the video domains and shows that an even higher masking ratio leads to strong performance for activity recognition tasks. V-JEPA~\cite{bardes2024revisiting} explores feature prediction as an objective for unsupervised learning from video and achieves state-of-the-art results on activity recognition task in the Something-Something V2 dataset\cite{goyal2017something}. However, the usefulness of these representations for physical dynamics understanding remains unexplored. Furthermore, these models are mostly used in a transfer learning or fine-tuning paradigm, which requires ground-truth annotations. It remains unclear whether they can be prompted to extract meaningful structures without additional training on annotated data.


\section{Method}

\begin{figure*}[t]
    \centering
    \includegraphics[width=0.98\linewidth]{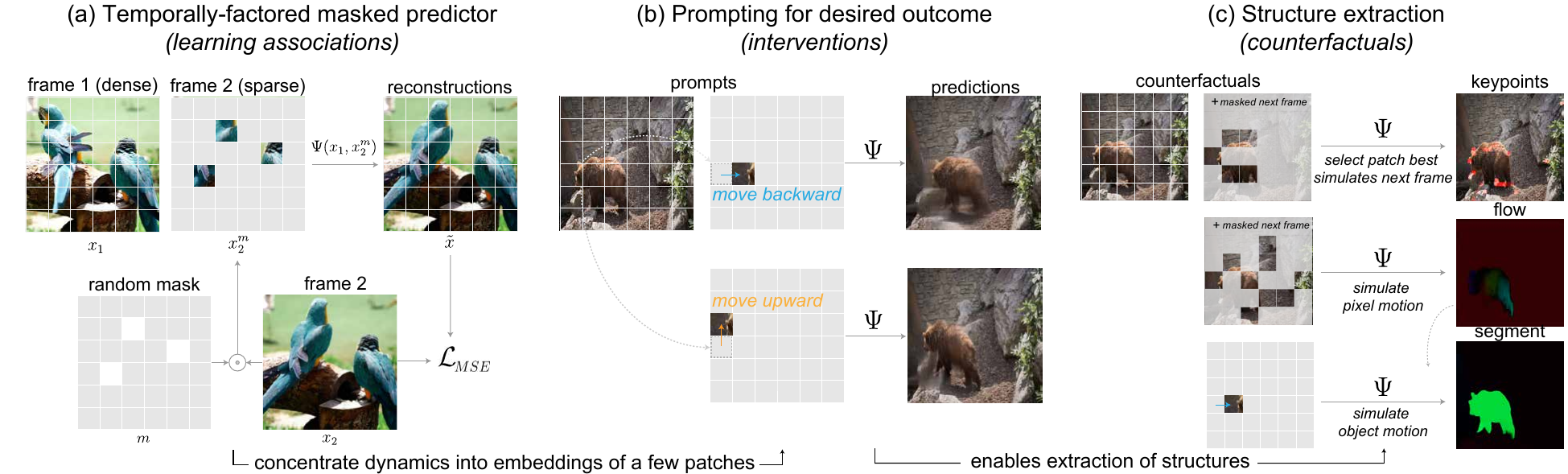}
    \caption{\textbf{Climbing the Ladder of Causation with the CWM framework: (a) Temporally-factored masked predictor for association learning.} Given a frame pair input, the predictor takes in dense visible patches from the first frame and only a sparse subset of patches from the second frame as inputs, and learns to predict the masked patches. This policy encourages the model to concentrate scene dynamics into embeddings of a few patches. \textbf{(b) Prompting as interventations.} As a result of the temporally-factored masking, we can intervene by modifying one or a few visual patches in the prompt and steer the outcome of the predictor. \textbf{(c) Structure extraction using counterfactuals.} Multiple vision structures can be extracted by comparing the results of interventions to alternative futures (e.g. observed ground truth or observed predictions).}
    \label{fig:1}
\end{figure*}

We discuss in generality the three concepts of CWM by climbing Pearl’s Ladder of Causation \cite{goldberg2019book}: (1) temporally-factored masked predictor for learning associations, (2) prompting as interventions and (3) structure extractions using counterfactuals.  We will discuss the application of CWM to physical dynamics understanding in Section \ref{experiment}.


\subsection{Temporally-factored masked predictor for learning associations}
\label{section_3_1}
\subsubsection{Masked predictor} Following MAE \cite{he2022masked} and VideoMAE \cite{tong2022videomae,feichtenhofer2022masked}, we train an encoder-decoder architecture to reconstruct masked observations of video frames. The input video frames are first divided into non-overlapping spatiotemporal square patches. Then a subset of the patches is masked, and only the remaining visible patches are passed as inputs into the encoder. Finally, the embeded tokens from the encoder and learnable mask tokens, with added positional embedding on all the tokens, are passed as inputs into a shallow decoder to reconstruct the masked patches. The predictor is trained with the mean squared error (MSE) loss between the reconstructed patches and the original masked patches. The predictor learns the associations between spatiotemporal patches of observed video inputs.

\subsubsection{Temporally-factored masking} Unlike VideoMAE \cite{tong2022videomae,feichtenhofer2022masked}, which randomly samples ``tubes'' or ``cubes'' of spatiotemporal patches to be masked, we use a temporally-factored masking policy for video inputs. Without loss of generality, we discuss the masking policy with a frame pair $x_1, x_2 \!\in\! \mathbb{R}^{3 \times H \times W}$ as input. Given the input frame pair, we train a predictor $\Psi$:
\begin{equation}
    \Psi(x_1^\alpha, x_2^\beta) = \tilde{x}_2
\end{equation}
which takes in first frame $x_1$ and second frame $x_2$ with masking ratio $\alpha, \beta \in [0,1]$. The predictor $\Psi$ predicts the masked patches of $x_2$, and minimizes the MSE loss between the reconstructed patches $\tilde{x}_2$ and the masked patches of $x_2$. Figure \ref{fig:1}\textcolor{red}{a} illustrates this masking policy.

Here, we set the masking ratio $\alpha$ to 0 and $\beta$ to 0.90, a highly asymmetric masking policy. As a result of this high masking ratio, the predictor $\Psi$ learns to complete the second frame given only a few patches of it, along with the fully visible first frame; the only way it can do this is by inferring scene transformations from a few second-frame patches, then applying these transformations to the first-frame patches to complete the second frame \cite{bear2023unifying}. This implies that the predictor learns to concentrate transformations between frame pairs into the embeddings of a few visible patches. Consequently, modifying the contents of a few patches, which represent transformations, can exert meaningful control over the next-frame predictions.

\subsection{Prompting as interventions}
\label{section_3_2}
\begin{figure*}[t]
    \centering
    \includegraphics[width=0.9\linewidth]{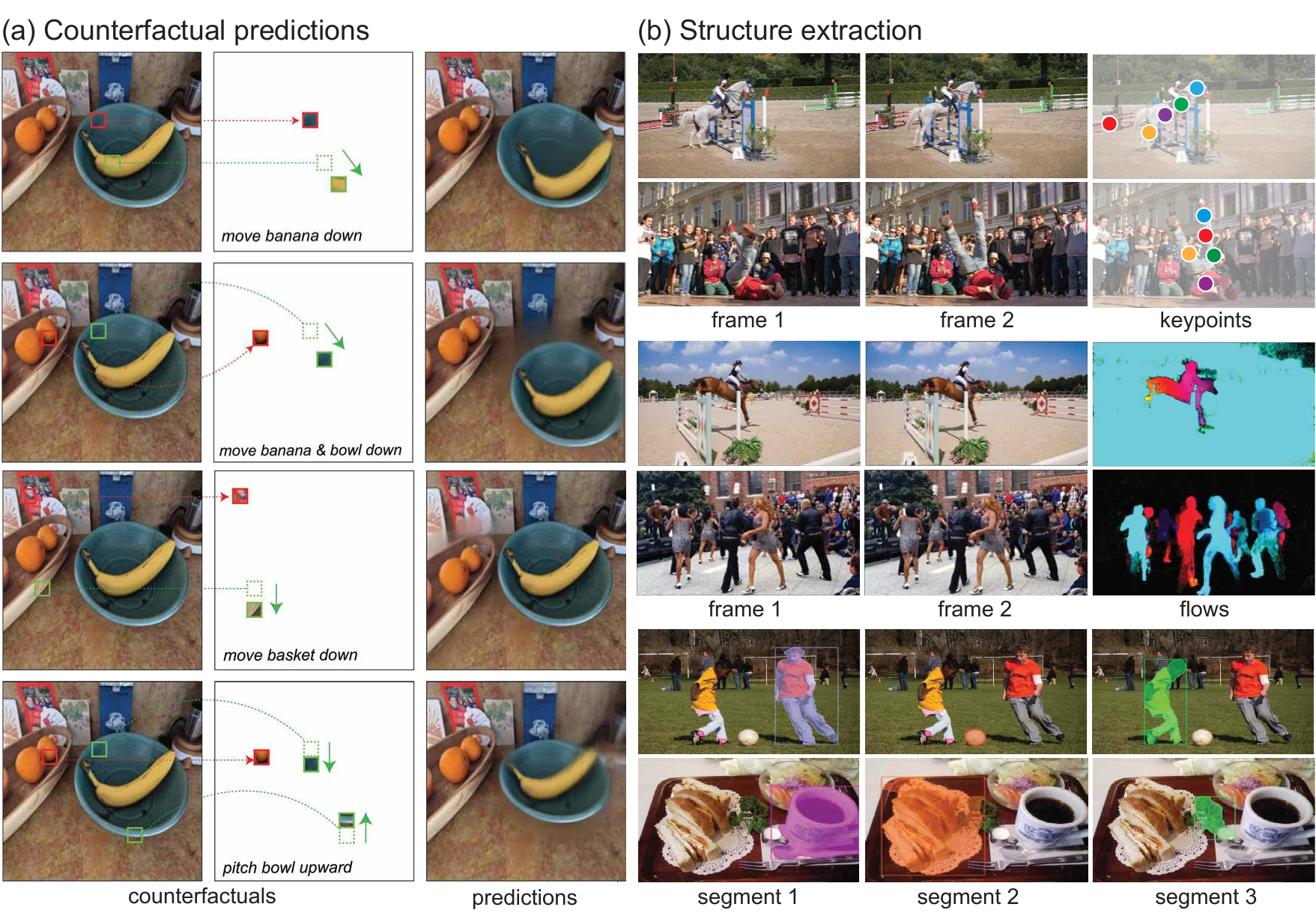}
    \caption{\textbf{Counterfactual predictions and structure extraction}. \textbf{(a) Counterfactual predictions}. A small number of visual patches exert meaningful control of scene dynamics. Each panel shows a prompt consisting of the input image (left), a few patches copied from the input image (middle), and the resulting predictions (right). A red patch is copied into the same location as its source, simulating the appearance of holding an object fixed.  A green patch is copied into a different location at an offset from the source location, simulating the appearance of an apparent object motion. \textbf{(b) Structure extraction} for keypoints, flows, and segments}
    \label{fig:2}
\end{figure*}

With a pre-trained predictor, at inference time we can replace empirical data observations with interventions intended to produce some desired outcome\cite{goldberg2019book}. As a result of the temporally-factored masking policy, we can modify the original inputs at a few patch locations to generate alternative outcomes using the predictor. To formalize the procedure of intervention, we first define a prompt $p$ as a set of video frames that is given as input to the predictor:
\begin{equation}
    p = \{x_1, x_2 \ | \, x_1, x_2 \!\in\! \mathbb{R}^{3 \times H \times W}\}
\end{equation}
\noindent where $x_2$ has a small number of visible patches that specify scene transformations. An intervention $\bar{p}$ is defined as an input to the predictor that has been modified from the initial prompt $p$. We use two basic types of interventions: (a) appearance prompts, which involve modifications to the first frame $x_1$, and (b) motion prompts, which involve modifications to the second frame $x_2$. Given a intervention $\bar{p}$, the associated prediction is the outcome of the predictor $\Psi(\bar{p})$ \cite{bear2023unifying}. Figure \ref{fig:2}\textcolor{red}{a} shows the predictions for a series of motion prompts. These prompts use a single image, $x_1$ and construct $x_2$ by revealing only a few patches in the input image and translating them by a small offset.

\subsection{Structure extraction using counterfactuals}
\label{section_3_3}
\label{sec:programs_struct}

The observation of the previous section shows how it is possible to generate counterfactual object motion by modifying the positions of a small number of patches. Next, we discuss how different structure extractions can be specified as counterfactuals~\cite{goldberg2019book}  by comparing the outcomes of the interventions with alternative futures (e.g. observed ground-truth data or predictions).

\subsubsection{Keypoints} have been previously defined by manual category-specific annotations~\cite{everingham2010pascal,lin2014microsoft,you2020keypointnet}. CWM provides a general category-agnostic definition of keypoints as patch locations in $x_2$ that, when revealed to the predictor, yield the lowest error in the reconstruction $\Psi(p)$ \cite{bear2023unifying}. Let $\mathcal{I}$ be a set of patch locations of an image. The set of keypoints is defined as:
\begin{equation}
\begin{gathered}
K(x_1,x_2, n) = \argmin_{k \tiny\subset \mathcal{I}, \, |k|=n} \mathcal{L}(\Psi(p), \, \Psi(\bar{p}))\, \\
\text{where} \quad \bar{p} = \{x_1, x_2^m \, | \, x_2^m \, \text{is visible at} \, k \}
\end{gathered}
\end{equation}
Here, the intervention $\bar{p}$ is the modification of the original input $p=\{x_1, x_2\}$, where the second frame $x_2^m$ is masked everywhere except at keypoint locations. This construction defines a set of dynamical RGB keypoints on $x_2$. For large values of $n$, this is in general an intractable optimization problem. In practice, we thus first start with an empty set and add keypoints one at a time to greedily reduce the reconstruction error until $n$ keypoints have been obtained. We show examples of extracted keypoints in the top panel of Figure \ref{fig:2}\textcolor{red}{b}.





\subsubsection{Optical flow}

is the task of estimating per-pixel motion between video frames\cite{teed2020raft}. To estimate per-pixel motion, we introduce an appearance intervention that adds a small perturbation to the pixel in the first frame. We can estimate the pixel motion by localizing the perturbation response in $\Psi(\bar{p})$ \cite{bear2023unifying}. 

More specifically, given a prompt $p=\{x_1,x_2^\beta\}$ and a pixel location $(i,j)$, we construct an intervention $\bar{p}=\{x_1{+}\delta_{ij}, x^\beta_2\}$, which adds a small perturbation $\delta_{ij}$ to the first frame at the pixel location. This creates a perturbed first frame by modifying its appearance at a pixel location. For this reason, we call this an appearance intervention. With a perturbed first frame, the predictor propagates the perturbation in the next frame, under the original scene transformations specified by $x_2^\beta$. The corresponding pixel location in the next frame can be localized by finding the peak of the perturbation response. The perturbation response in the next frame can be computed as the absolute difference between the counterfactual prediction $\Psi(\bar{p})$ and the observed prediction $\Psi(p)$. Then, we locate the peak of the perturbation response by taking an argmax over the set of patch locations $\mathcal{I}$. The flow at pixel location $(i,j)$ is then defined as the spatial displacement between $(i,j)$ and the peak of perturbation response:

\begin{equation}
    F_{i,j}(x_1, x_2) = \argmax_\mathcal{I} \, |\,\Psi(\bar{p}) - \Psi(p)\,| - (i, j)
\end{equation}

This algorithm is simple and often effective, as shown in the middle panel of Figure \ref{fig:2}\textcolor{red}{b}, but it might fail in two ways. First, one of the revealed patches in $x_2^\beta$ may cover the place where the perturbation at location $(i,j)$ is expected to move. This can be remedied by running the above procedure for multiple random choices of $x_2^\beta$ and taking their average perturbation responses \cite{bear2023unifying}.


A second potential failure mode is that the intervention $\bar{p}$ might be out of distribution for $\Psi$, which could happen when the perturbation $\delta_{ij}$ is too large \cite{bear2023unifying}. On the other hand, if the perturbation is too small, it might not be detected and moved accurately.
This can be naturally addressed by using infinitesimal perturbations. We normalize the magnitude of the perturbation response by the magnitude of the perturbation as the limit goes to zero. This is exactly the derivative of the $\Psi$ at location $(i,j)$:

\begin{align}
\lim_{\delta_{ij}\rightarrow 0} \frac{|\Psi(\bar{p}) - \Psi(p)|}{|\delta_{ij}|} \,\, = \nabla_{x}\Psi \bigg|_{(i,j)} 
\end{align}

To simultaneously estimate optical flow at all locations of an input frame, we can compute the Jacobian of $\Psi$. This is a tensorial operation that can be computed once at all pixels using PyTorch autograd \cite{torch.autograd}. We describe more details about the procedures of extracting flow in the supplementary material.





\subsubsection{Segmentation} is defined as a collection of physical stuff that moves together under the application of everyday physical actions \cite{chen2022unsupervised}. This is inspired by the notion of Spelke object in infant object recognition: infants tend to group scene elements that move together as a single object \cite{spelke1990principles}. CWM extracts segmentation of objects by motion interventions, which simulate object motion at a pixel location, followed by grouping parts of the image that move together.


Given a single image $x$ as input, we define an intervention $\bar{p} = \{x, \bar{x}^m\}$. These prompts produces the second frame $\bar{x}^m$ by revealing only a few patches in the input image and translating those patches by a small offset. With a temporally-factored masked predictor, moving a few patches in the prompt will cause the entire object to move in the resultant counterfactual predictions $\Psi(\bar{p})$. Segments can be extracted by thresholding the flows between the input image $x$ and $\Psi(\bar{p})$:

\begin{equation}
    S(x) = F(x, \,\Psi(\bar{p})) > 0
\end{equation}


Once a segment is extracted, we iterate the procedure above to refine the segment by revealing more patches within the segment region into $\bar{x}^m$ and translating patches in the same direction. We set the number of iterations as 3. To automatically discover multiple objects in a single image, we iteratively extract segments at pixel locations that are not part of a discovered object. Once an object segment is discovered, we reveal patches that are not within the segment region and repeat the procedure to discover the next object. We show examples of extracted segments in the bottom panel of Figure \ref{fig:2}\textcolor{red}{b}. We discuss more details in the supplementary material.

\section{Experiments}
\label{experiment}

Section~\ref{sec:Physion} first investigates the usefulness of the extracted structures for downstream physical dynamics understanding tasks. Section ~\ref{sec:analysis_struct} evaluates the quality of counterfactual motion predictions and extracted visual structures on real-world datasets. Section~\ref{sec:ablations} discusses ablations studies on the CWM design.

\subsection{Physical Dynamics Understanding}
\label{sec:Physion}

\noindent \textbf{Physion benchmark} consists of realistic simulations of diverse physical scenarios where objects are manipulated in a variety of configurations to test different types of physical reasoning such as stability, rolling motion, object linkage, etc. We use the latest version of Physion~\cite{bear2021physion}, referred to as Physion v1.5\footnote{ https://physion-benchmark.github.io}, which has improved rendering quality and more physically plausible simulations. 

In the ideal scenario, we would evaluate CWM on a real-world physics-understanding benchmark, but such benchmarks are not available. Recent works have shown that simulated data can be highly valuable \cite{liu2023zero,teed2020raft,zheng2023pointodyssey}. Physion is a challenging benchmark as it contains diverse physical phenomena, object dynamics and realistic 3D simulations. This makes it a preferable choice when compared to other benchmarks such as ShapeStacks~\cite{groth2018shapestacks} and IntPhys~\cite{riochet2021intphys} which contain very limited object dynamics, or Phyre~\cite{bakhtin2019phyre} which only operates in 2D environments. Existing video models still significantly lag behind human performance on the Physion benchmark\cite{bear2021physion}. Moreover, the CWM model is trained on real-world videos from Kinetics-400 dataset\cite{kay2017kinetics} and tested on Physion, and is thus a strong generalization test.

\begin{figure}[t]
    \centering
    \includegraphics[width=0.95\linewidth]{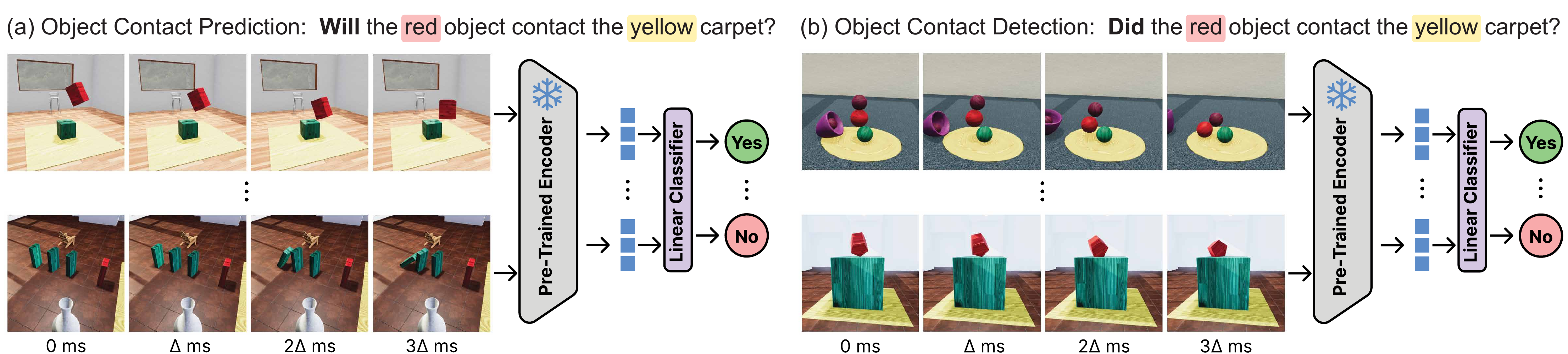}
    \caption{\textbf{Physion v1.5 evaluation protocol.} We evaluate on two physical dynamics understanding tasks -- \textbf{(a) Object contact prediction} where the model is asked to predict contact events in the future and \textbf{(b) Object contact detection} where the model is asked to reason about contact events that occur in the observed video stimulus. The objects of interest for which we want to ask the contact question are rendered with red and yellow texture to cue the model. }
    \label{fig:physion_eval}
\end{figure}

The benchmark consists of two tasks: (a) \textit{Object contact prediction} (OCP), which tests the model's ability to \textit{predict} whether two objects \textit{will} contact at some point in the future given a context video, and (b) \textit{Object contact detection} (OCD) which tests the model's ability to \textit{detect} if two objects have already come into contact in the observed video. The video stimuli are generated in such a way that the model needs to have an understanding of the physical dynamics in order to answer the contact-related question correctly. Figure~\ref{fig:physion_eval} shows  example stimuli for the two tasks. For both tasks, the two objects of interest are rendered with red and yellow texture to cue the model. 



\textbf{Evaluation protocol} 
We follow the three-step evaluation protocol of the Physion benchmark~\cite{bear2021physion}. First, we extract features from the last layer of a frozen pre-trained encoder on a training set of 5,600 videos for OCP and OCD tasks, respectively. For image-based methods, features are extracted from 4 frames that are 150 ms apart. For video-based methods, the input frames are fed to the model at the specific frame rate used during their training. Second, the extracted features are used to train a logistic regression model to predict the contact label for the given video stimulus. Lastly, the trained classifier is evaluated on a test set of 1,000 videos across different physical scenarios.

\textbf{Baseline methods}
We compare CWM with five classes of baseline approaches: (a) video prediction models including MCVD \cite{voleti2022mcvd}, R3M \cite{nair2022r3m}, FitVid \cite{babaeizadeh2021fitvid}, and TECO \cite{yan2023temporally}, (b) self-supervised representation learning methods on images including DINO \cite{caron2021emerging}, DINOv2 \cite{oquab2023dinov2}, and MAE \cite{he2022masked}, (c) self-supervised representation learning methods on videos including VideoMAE \cite{tong2022videomae}, VideoMAEv2 \cite{wang2023videomae} and the recent state-of-the-art method V-JEPA~\cite{bardes2024revisiting} (d) vision-language models like GPT4-V \cite{OpenAI2023GPT4Vision} and lastly (e) ground truth 3D particle-based dynamics prediction models such as SGNN~\cite{han2022learning}. 

\begin{table}[t!]
  \centering
  \scriptsize
  \caption{\textbf{State-of-the-art accuracy on Physion v1.5}. We compare CWM to five classes of baseline methods across different architectures on the OCP and OCD tasks. We evaluate CWM with both features and extracted structures and find that it achieves state-of-the-art performance on these tasks. Original VideoMAE \cite{tong2022videomae} uses 16 input frames and a patch size of 16. We trained VideoMAE* with 3 input frames, a patch size of 8, and include extracted vision structures from the model for a strictly fair comparison with CWM.}
  
  \begin{tabular*}{.85\linewidth}{@{\extracolsep{\fill}}llllcc}
    \toprule
    method & training data & arch & param & OCP\,$\uparrow$ & OCD\,$\uparrow$  \\
    \midrule
    \multicolumn{6}{l}{\textcolor{gray}{\textit{supervised ground truth 3D particle-based model}}} \\
    \midrule
    \textcolor{gray}{SGNN} & \textcolor{gray}{Physion v1.5} & \textcolor{gray}{GNN} & \textcolor{gray}{23 M} & \textcolor{gray}{76.4} & \textcolor{gray}{98.8}  \\
    \midrule
    \multicolumn{6}{l}{\textit{video prediction models}} \\
    \midrule
    MCVD \cite{voleti2022mcvd} & K400+Ego4D  & UNet & 251 M & 63.4 & 80.8 \\ 
    R3M \cite{nair2022r3m} & K400+Ego4D  & Res50 & 38 M & 67.6 & 78.1 \\
    FitVid \cite{babaeizadeh2021fitvid} & K400+Ego4D  &VAE & 303 M & 64.3 & 59.5 \\
    TECO \cite{yan2023temporally} & K600  & vq-gan & 160 M & 69.3 & 80.9 \\
    \midrule
    \multicolumn{6}{l}{\textit{self-supervised image representation models}} \\
     \midrule
    DINO \cite{caron2021emerging} & IN-1K  & ViT-B & 86M & 72.1 & 85.4 \\
    DINOv2 \cite{oquab2023dinov2} & LVD-142M  & ViT-B & 86 M & 72.2 & 87.1 \\
    DINOv2 \cite{oquab2023dinov2} & LVD-142M  & ViT-L & 304 M & 72.2 & 85.5 \\
    DINOv2 \cite{oquab2023dinov2} & LVD-142M  & ViT-g & 1.1 B & 72.7 & 84.6\\
    MAE \cite{he2022masked} & IN-1K  & ViT-B & 86 M & 72.6 & 81.6\\
    MAE \cite{he2022masked} & IN-1K  & ViT-L & 304 M & 71.6 & 82.3 \\
    MAE \cite{he2022masked} & IN-1K  & ViT-H & 632 M & 73.3 & 80.8 \\
    MAE \cite{he2022masked} & IN-4.5M  & ViT-B & 86 M & 72.1 & 81.7 \\
    MAE \cite{he2022masked} & IN-4.5M  & ViT-L & 304 M & 72.6 & 81.9 \\
    
    \midrule
    
    \multicolumn{6}{l}{\textit{self-supervised video representation models}} \\
     \midrule
    VideoMAE \cite{tong2022videomae} & K400  & ViT-B & 86 M & 72.1 & 85.7  \\
    VideoMAE* & K400  & ViT-B & 86 M & 73.2 & 86.2  \\
    VideoMAE \cite{tong2022videomae} & K400  & ViT-L & 304 M & 73.6 & 86.1 \\
    VideoMAE \cite{tong2022videomae} & K400  & ViT-H & 632 M & 73.5 & 87.5 \\
    VideoMAEv2 \cite{wang2023videomae} & U-Hybrid  & ViT-g & 1.1B & 72.2 & 85.0 \\
    V-JEPA \cite{bardes2024revisiting} & VideoMix2M & ViT-L & 304M & 73.4 & 87.0 \\
    \midrule
    \multicolumn{6}{l}{\textit{vision-language models}} \\
    \midrule
    GPT4-V \cite{OpenAI2023GPT4Vision}  & - & - & - & 52.9 & 54.7 \\
    \midrule
    \midrule
    CWM & K400 & ViT-B & 86 M & 75.9 & \textbf{89.1}\\
    CWM & K400  & ViT-L & 304 M & \textbf{76.1} & 88.7\\
    
    \bottomrule
  \end{tabular*}
  
  \label{tab:main_results}
\end{table}
\textbf{Results}
In Table~\ref{tab:main_results} we report results on the two Physion tasks for both CWM with ViT-B and ViT-L architectures and other baseline methods discussed above. We evaluate CWM with both features and extracted vision structures input to the linear classifier. We find that video prediction models (such as MCVD~\cite{voleti2022mcvd} and TECO~\cite{yan2023temporally}) perform poorly especially on the Physion tasks. Self-supervised image representation models on the other hand, are better but they saturate around 72\% and 87\% for OCP and OCD respectively with the ViT-B architecture. It is interesting to note that CWM outperforms methods such as DINOv2 ViT-g and MAE ViT-H which have 13 and 7 times more parameters. When scaled up to ViT-L, CWM achieves superior performance on OCP. 

We find that CWM exhibits superior performance compared to both VideoMAE~\cite{tong2022videomae} and VideoMAEv2~\cite{wang2023videomae}. To ensure a fair evaluation, we train a variant of VideoMAE, denoted as VideoMAE*, that matches CWM in terms of the number of frames and patch size, and include comparable structure extractions from the model for linear probing. Our findings indicate that CWM performs better than VideoMAE*. Furthermore, CWM surpasses the recently released V-JEPA~\cite{bardes2024revisiting}, a state-of-the-art model for video representation, despite being trained on a considerably smaller dataset. Furthermore, we find on OCP, CWM achieves a performance that closely approaches that of ground truth 3D particle-based simulation models (i.e SGNN~\cite{han2022learning}) learned on Physion, despite being trained on Kinetics-400~\cite{kay2017kinetics} -- a considerably different real world dataset. 

We also evaluate GPT4-V~\cite{OpenAI2023GPT4Vision} on Physion v1.5 tasks by providing it with a single composite image with a sequence of four video frames sampled at a gap of 150ms. The model is prompted with questions similar to those in Figure~\ref{fig:physion_eval} (see supplementary for more details about the specific prompts used). We find GPT4-V scores nearly at chance on OCP and slightly above chance on OCD, which highlights a considerable limitation in the ability of large-scale vision-language models to understand physical scene dynamics.



\subsection{Analysis of the extracted structures}
\label{sec:analysis_struct}
We analyze the quality of structures extracted by CWM. Although not all baseline methods can perform counterfactual predictions or structure extractions, we apply our procedures to the baseline methods for a fair comparison with CWM. We show CWM yields more meaningful predictions, keypoints, and flows than VideoMAE, enabled by the temporally-factored masking policy. We also show that the quality of segments extracted by CWM is close to the state-of-the-art method CutLER\cite{wang2023cut}, which extracts segments from DINO \cite{caron2021emerging}.

\begin{figure*}[t]
    \centering
    \includegraphics[width=0.9\linewidth]{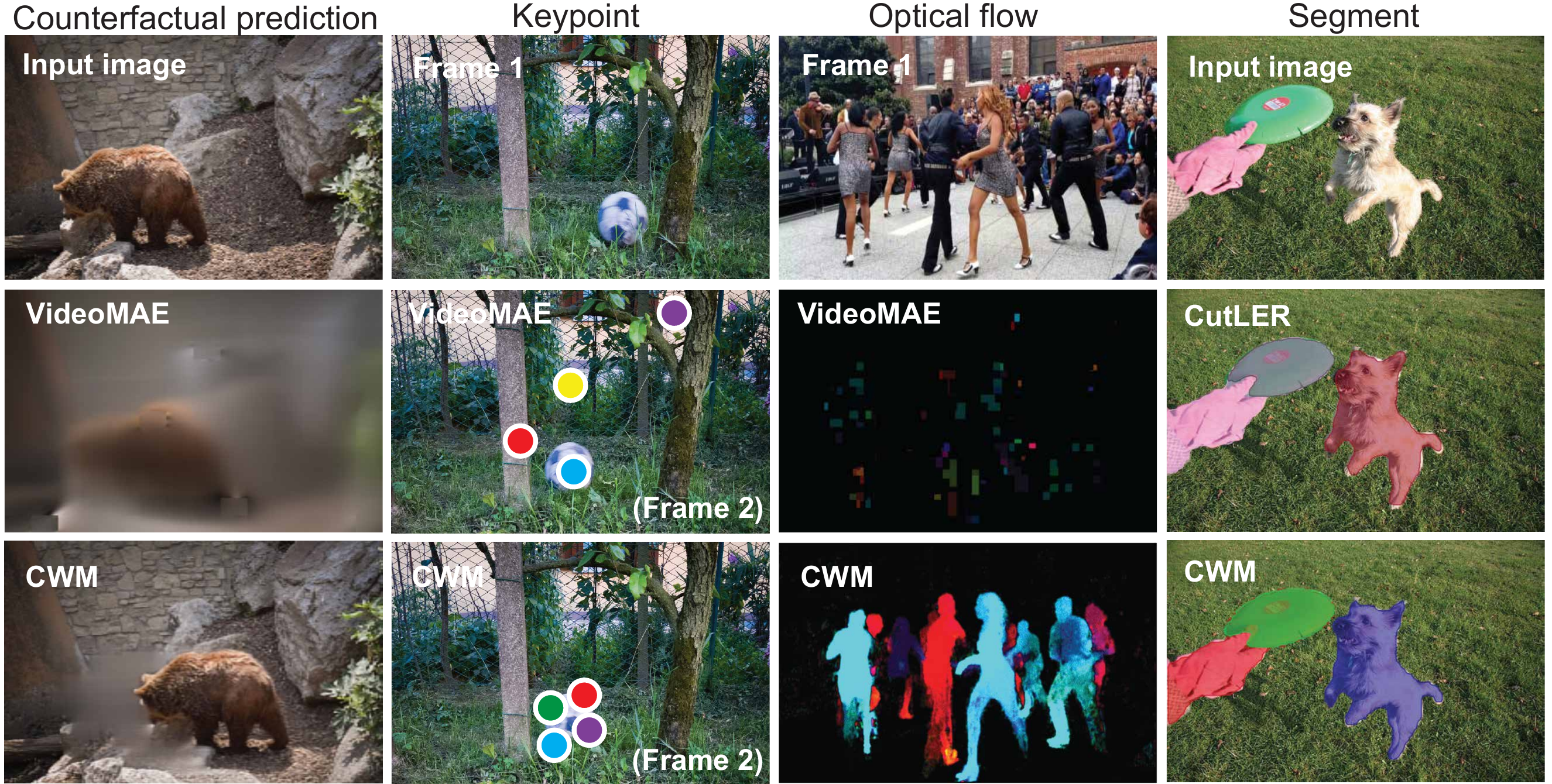}
    \caption{\textbf{Qualitative comparison of counterfactual motion prediction and structure extraction on real-world datasets}. We find that when we apply our extraction procedures described in Section.~\ref{sec:programs_struct} on VideoMAE, the model fails to generate counterfactual motion and extracts less meaningful structures than CWM. Segments cannot be extracted from VideoMAE due to the failure of counterfactual predictions, and hence not shown in this comparison. This shows the importance of the temporally-factored masking policy during pre-training.}
    \label{fig:structure_comparison}
\end{figure*}
\begin{table}[t]
    \centering
    \scriptsize
    \setlength{\tabcolsep}{4pt}
    \label{table:combined}
    \caption{\textbf{Quantitative comparison of counterfactual motions, flow and segment extraction on real-world datasets.} In (a) we compare to VideoMAE on counterfactual motions and flow. For a strictly fair comparison, we also evaluate VideoMAE* which we trained with the same patch size and number of frames as CWM. In (b) we compare CWM to CutLER \cite{wang2023cut}, which extracts segmentations from DINO\cite{caron2021emerging}, and FreeSOLO\cite{wang2022freesolo} on the quality of segmentations.}
    \begin{subtable}{.55\textwidth}
        \centering
        \caption{Counterfactual motion (CM) and Flow}
        \vspace{-0.3cm}
        \begin{tabular}{lcc}
            \toprule
            Methods & CM (FID $\downarrow$) & Flow (F1 $\downarrow$) \\
            \midrule
            VideoMAE~\cite{tong2022videomae} & 213.4 & 56.3 \\
            VideoMAE* & 166.3 & 54.9 \\
            CWM & \textbf{25.4} & \textbf{46.8} \\
            \bottomrule
        \end{tabular}
        \label{subtable:counterfactual_flow}
    \end{subtable}
    \hfill 
    \begin{subtable}{.4\textwidth}
        \centering
        \caption{Segments extraction}
        \vspace{-0.3cm}
        \begin{tabular}{lc}
            \toprule
            Methods & Segment (AP $\uparrow$) \\
            \midrule
            FreeSOLO \cite{wang2022freesolo} & 4.3 \\
            CutLER \cite{wang2023cut} & \textbf{8.4} \\
            CWM & 8.2 \\
            \bottomrule
        \end{tabular}
        \label{subtable:segment_extraction}
    \end{subtable}

\end{table}

\textbf{Counterfactual prediction} We compare CWM and VideoMAE on the quality of counterfactual predictions in Table~\ref{subtable:counterfactual_flow} and Figure \ref{fig:structure_comparison}. We generate counterfactual motions using input images from the DAVIS dataset~\cite{davis2017}. The quality of generation is measured by the Fr\'echet Inception Distance (FID)~\cite{heusel2017gans}. CWM significantly outperforms VideoMAE. For a strictly fair comparison, we train another VideoMAE model (referred to as VideoMAE*) with the same number of frames and patch size as CWM. Although the model achieves a slightly lower FID relative to VideoMAE, the reconstructions are still quite blurry without accurate object motions. This illustrates the importance of temporally-factored masking in generating plausible counterfactual predictions.  

\textbf{Keypoints} Existing keypoint datasets are generally created with manually specified templates for certain object categories~\cite{lin2014microsoft, you2020keypointnet,everingham2010pascal}. Therefore, these datasets do not provide suitable quantitative evaluations of CWM keypoint, which are category-agnostic. Figure \ref{fig:structure_comparison} shows that CWM can extract more meaningful dynamic keypoints as compared to VideoMAE. 



\textbf{Optical flow} We evaluate the quality of optical flows on the SPRING benchmark \cite{Mehl2023_Spring} using the F1 metric \cite{geiger2013vision}. We find that CWM is better compared to both VideoMAE and VideoMAE* (See Table.~\ref{subtable:counterfactual_flow}). This is also supported by the qualitative results shown in Figure.~\ref{fig:structure_comparison}. We include more qualitative comparisons and additional implementation details in the supplementary.

\label{sec:seg_exp}
\textbf{Segments} We extract segments on images from COCO~\texttt{train2017}~\cite{lin2014microsoft} using CWM.  We follow the same procedures in CutLER~\cite{wang2023cut} to learn a detector using the extracted segments as self-supervision. We train CutLER on COCO training images for a fair comparison. We compare CWM with FreeSOLO~\cite{wang2022freesolo} and CutLER~\cite{wang2023cut} in Table \ref{subtable:segment_extraction} and Figure \ref{fig:structure_comparison}. CWM outperforms FreeSOLO \cite{wang2022freesolo} significantly and attains similar performance to the current state-of-the-art approach CutLER \cite{wang2023cut}. Although Spelke objects are segment-like structures, the definition of Spelke objects is not exactly aligned with the definition of instance segmentations in the COCO datasets.

\subsection{Ablation studies}
\label{sec:ablations}
We ablate the CWM design with the default backbone of ViT-B. Each ablated model is trained for 800 epochs on the Kinetics-400 dataset. Results of the ablation study are reported in Table.~\ref{table:ablation}.


\textbf{Vision structures} We study the importance of each visual structure in understanding dynamics. Adding patch features at keypoint locations improves the OCP accuracy from 73.6\% to 74.4\%. Enriching these patch features with optical flow patches further improves the accuracy to 75.5\%. Finally, including segments achieves a score of 75.9\%. 

\textbf{Training schedule} We find that a model trained with a longer training schedule of 1600 epochs achieves an OCP score of 75.9\% -- a relatively small improvement over an 800 epoch trained model (75.4\%).

\textbf{Masking Policy} We study the importance of temporal factoring by training a model with a random tube masking strategy, which was originally proposed in VideoMAE~\cite{tong2022videomae}. The temporally-factored mask policy is essential for extraction of meaningful vision structures, improving the OCP accuracy improves from 73.2\% with tube masking to 75.9\% with temporally-factored masking. 


\textbf{Mask ratio} We observe that a high ratio on the last frame (90\%) during model training achieves good performance on both the OCP and OCD tasks. This trend aligns with our aforementioned hypothesis that the dynamics between frame pairs at a short timescale has a low-dimensional causal structure, which can be concentrated into a small number of tokens.



\begin{table}[t] 
    \centering
    \scriptsize
    
    \centering
    \caption{\textbf{CWM ablation studies.} The best setting is shown in the first row. We investigate the importance of different vision structures, masking policy, training epochs, masking ratio, context frames and patch size.}
    \begin{tabular}{p{1.5cm}cccccccccccc}
    \toprule
    \multirow{2}{*}{Ablations} & \multirow{2}{*}{\parbox{1.5cm}{\centering temporal \\factoring}} & \multicolumn{4}{c}{Input to the classifier} & \multirow{2}{*}{\parbox{0.8cm}{\centering mask\\ratio}} & \multirow{2}{*}{\parbox{0.8cm}{\centering patch\\size}} & \multirow{2}{*}{\parbox{1cm}{\centering context\\frames}} & \multirow{2}{*}{\parbox{1cm}{\centering training\\epochs}} & \multicolumn{2}{c}{Metrics} \\
    \cmidrule(lr){3-6}  \cmidrule(lr){11-12} 
    & & feat. & keyp. & flow & segm. & & & &  & OCP $\uparrow$ & OCD $\uparrow$ \\
    \midrule
    \multirow{1}{*}{Best setting} & \checkmark & \checkmark & \checkmark & \checkmark & \checkmark & 0.90 & 8 & 2 & 1600  &  \textbf{75.9} &  \textbf{89.1}  \\
    
    \midrule
    \midrule
    \multirow{3}{*}{Structures} & \checkmark & \checkmark & \checkmark & \checkmark & \xmark & 0.90 & 8 & 2 & 1600  & 75.5 & 88.5   \\
    & \checkmark & \checkmark & \checkmark & \xmark  & \xmark & 0.90 & 8 & 2 & 1600  & 74.4 & 89.1   \\
    & \checkmark & \checkmark & \xmark & \xmark & \xmark & 0.90 & 8 & 2 & 1600 &  73.6 & 89.1 \\
   
    \midrule
    \multirow{1}{*}{Training epochs} & \checkmark & \checkmark & \checkmark & \checkmark & \checkmark & 0.90 & 8 & 2 & 800 & 75.4 & 88.9 \\

    \midrule
    \multirow{1}{*}{Masking policy} & \xmark & \checkmark & \checkmark & \checkmark & \checkmark & 0.90 & 8 & 2 & 800 & 73.2 & 86.2 \\
    
    \midrule
    \multirow{3}{*}{Masking ratio} & \checkmark & \checkmark & \checkmark & \checkmark & \checkmark & 0.85 & 8 & 2 & 800 & 75.0 & 88.9 \\
    & \checkmark & \checkmark & \checkmark & \checkmark & \checkmark & 0.95 & 8 & 2 & 800 & 74.6 & 88.3 \\
    & \checkmark & \checkmark & \checkmark & \checkmark & \checkmark & 0.99 & 8 & 2 & 800 & 72.5 & 86.6 \\
    \midrule
    \multirow{2}{*}{Context frames} & \checkmark  & \checkmark & \checkmark & \checkmark & \checkmark & 0.90 & 8 & 1 & 800 & 71.0 & 85.2 \\
    & \checkmark  & \checkmark & \checkmark & \checkmark & \checkmark & 0.90 & 8 & 4 & 800 & 68.5 & 79.9\\
    \midrule
    \multirow{1}{*}{Patch size} & \checkmark  & \checkmark & \checkmark & \checkmark & \checkmark & 0.90 & 16 & 2 & 800 & 74.2 & 88.8 \\

    \bottomrule
    \end{tabular}
    \label{tab:intermediates_ablation}
    
    \label{table:ablation}
\end{table}

\textbf{Context length} We compare the performance of CWM with different numbers of context frames. CWM with 2 context frames during pre-training performs better as compared to using 1 context frame. However, including 4 context frames degrades the performance. 

\textbf{Patch size} Our analysis indicates that the patch size used for training the model can influence the performance; specifically, a patch size of 8 yields a superior OCP accuracy of 75.9\%, compared to a patch size of 16, which results in a lower accuracy of 74.2\%.



\section{Conclusion}

In this work, we show that a simple temporally-factored masking policy during pre-training enables powerful prompting abilities. As a result, we can use counterfactual prompts and their associated predictions to extract vision structures, which abstract away irrelevant details and thus end up being useful for improved dynamics understanding.  As compared to random masking, temporally-factored masking policy allows more meaningful and useful structures to be extracted from the pre-trained predictor. CWM achieves state-of-the-art results on the challenging Physion benchmark as compared to previous self-supervised methods, approaching the performance of the best supervised methods in terms of object contact prediction accuracy.

\clearpage
\subsubsection{Acknowledgements} 
This work was supported by the following awards: To D.L.K.Y.: Simons Foundation grant 543061, National Science Foundation CAREER grant 1844724, Office of Naval Research grant S5122, ONR MURI 00010802 and ONR MURI S5847. We also thank the Google TPU Research Cloud team for computing support.

%
%
\bibliographystyle{splncs04}
\bibliography{main}

\clearpage

\startcontents

\setcounter{section}{0}

\printcontents{}{1}{\section*{Supplementary Material}\setcounter{tocdepth}{2}}

\phantomsection
\section{CWM pre-training}

\label{figmention:figure_1}
\subsection{Architecture details} 
\begin{figure}[t]
    \centering
    \includegraphics[width=1.0\linewidth]{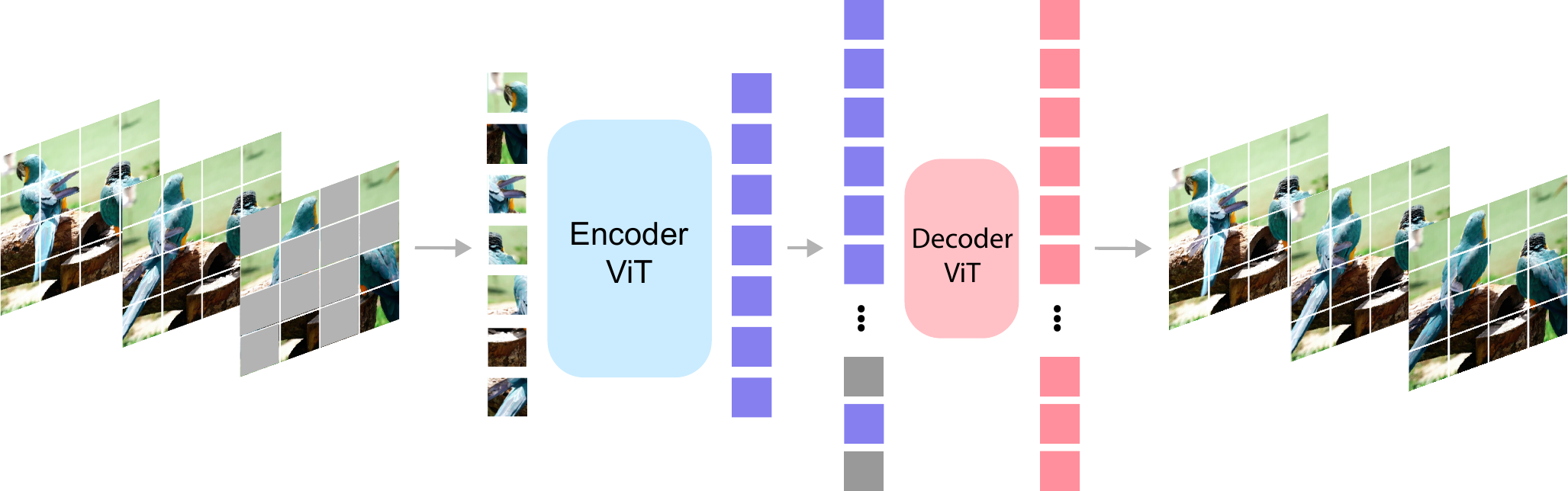}
    \caption{\textbf{Architecture of the masked predictor $\Psi$ in the CWM framework}.}
    \label{fig:arch}
    \vspace{-0.1cm}
    
\end{figure}

Figure \ref{fig:arch} \hypertarget{figure_1} provides an overview of the predictor architecture. The input video is first divided into non-overlapping spatiotemporal patches of size $8\times8$. Then a subset of patches is masked, and only the remaining visible patches are passed as inputs into the transformer encoder. We follow the standard ViT architecture. Following MAE \cite{he2022masked}, each transformer block in the ViT consists of a multi-head self-attention block and an MLP block, both having LayerNorm (LN). The CWM encoder and decoder have different widths, which are matched by a linear projection after the encoder \cite{he2022masked}. Finally, the embedded tokens from the encoder and learnable mask tokens are passed as inputs into a shallow decoder to reconstruct the masked patches. Each spatiotemporal patch has a unique sine-cosine positional embedding. Position embeddings are added to both the encoder and decoder inputs. CWM does not use relative position or layer scaling \cite{bao2021beit, he2022masked}.

\subsection{Implementation details}

While the discussion about CWM has for simplicity of presentation assumed a frame pair as input, for physical prediction problems it is natural to have an additional context frame to allow object initial velocities to be well-defined. More specifically, given 3 consecutive video frames at 150 ms apart during training, we provide full visibility to the first two context frames and only mask the last frame. In common situations where there is no motion in the first two frames, the three-frame model will recover what a two-frame model would have learned. When there is motion, the three-frame model will additionally learn acceleration, which is essential for physical predictions. For extracting keypoint and flow which only require 2 frames as input, we repeat the first frame twice so that the total input length is 3 frames. For extracting segmentation, we are given a single input frame, which we repeat twice and simulate object motions onto the third frame to compute segments. 

CWM uses the standard ViT-B and ViT-L architectures with a patch size of 8, which allows structure extraction at a higher resolution. We pre-train CWM on the Kinetics-400 dataset \cite{kay2017kinetics}, without requiring any specialized sparse operations or temporal downsampling.  It takes approximately 6 days to train 1600 epochs on a TPU v4-256 pod. 

\subsection{Default settings} We show the default pre-training settings in Table \ref{table:pre_training_setting}. CWM does not use color jittering, drop path, or gradient clip. Following ViT’s official code, xavier uniform is used to initialize all Transformer blocks. Learnable masked token is initialized as a zero tensor. Following MAE, we use the linear lr scaling rule: $lr = base\_lr\times batch\_size\,/\,256$ \cite{he2022masked}.

\begin{table}[h]
\centering

\caption{\textbf{Default pre-training setting of CWM}}
\begin{tabular}{l|l}
\toprule
\textbf{config}          &  \,value\\ \midrule
optimizer                & \,AdamW \cite{loshchilov2017decoupled}                         \\
base learning rate       & \,1.5e-4                        \\
weight decay             & \,0.05                          \\
optimizer momentum       & \,$\beta_1, \beta_2=0.9, 0.95$ \cite{chen2020generative}  \\
accumulative batch size         & \,4096                          \\
learning rate schedule   & \,cosine decay  \cite{loshchilov2016sgdr}                \\
warmup epochs \cite{goyal2017accurate}           & \,40                            \\
total epochs             & \,1600                          \\
flip augmentation        & \,no                            \\
augmentation             & \,MultiScaleCrop  \cite{wang2018temporal}       \\ \bottomrule
\end{tabular}
\label{table:pre_training_setting}
\vspace{-0.5cm}
\end{table}
\section{Structure extraction details and results}
In this section, we discuss implementation details of the counterfactual queries for extracting keypoints, optical flow, and segmentations. We also provide more qualitative results of each structure extracted by CWM. 

\subsection{Keypoint}

\subsubsection{Implementation details} CWM queries keypoints iteratively, starting with an intervention initialized as an initial empty mask and adding visible tokens one-by-one. Note that, whereas the counterfactual queries for optical flow and segmentation involve perturbing the visual input to the predictor $\Psi$, keypoints arise by varying the prediction model’s input mask. 

At each iteration, we compute the Mean-Squared-Error (MSE) between the next-frame predictions of $\Psi$ and the ground-truth next frame. We sort the MSE and select the top $k$ locations as candidate keypoints. $k$ is set as 4 by default. For each candidate keypoint, we add its patch content to the intervention and re-compute the MSE between the updated predictions of $\Psi$ and the ground-truth next frame. The candidate keypoint with the minimum MSE error, or equivalently maximum error reduction, is selected as the keypoint output at that iteration. The selected keypoint is added to the intervention and we repeat the procedures above to compute the location of the next keypoint. 

\label{figmention:keypoint}
\subsubsection{Additional qualitative results} 
\begin{figure*}
  \centering
  \includegraphics[width=\linewidth]{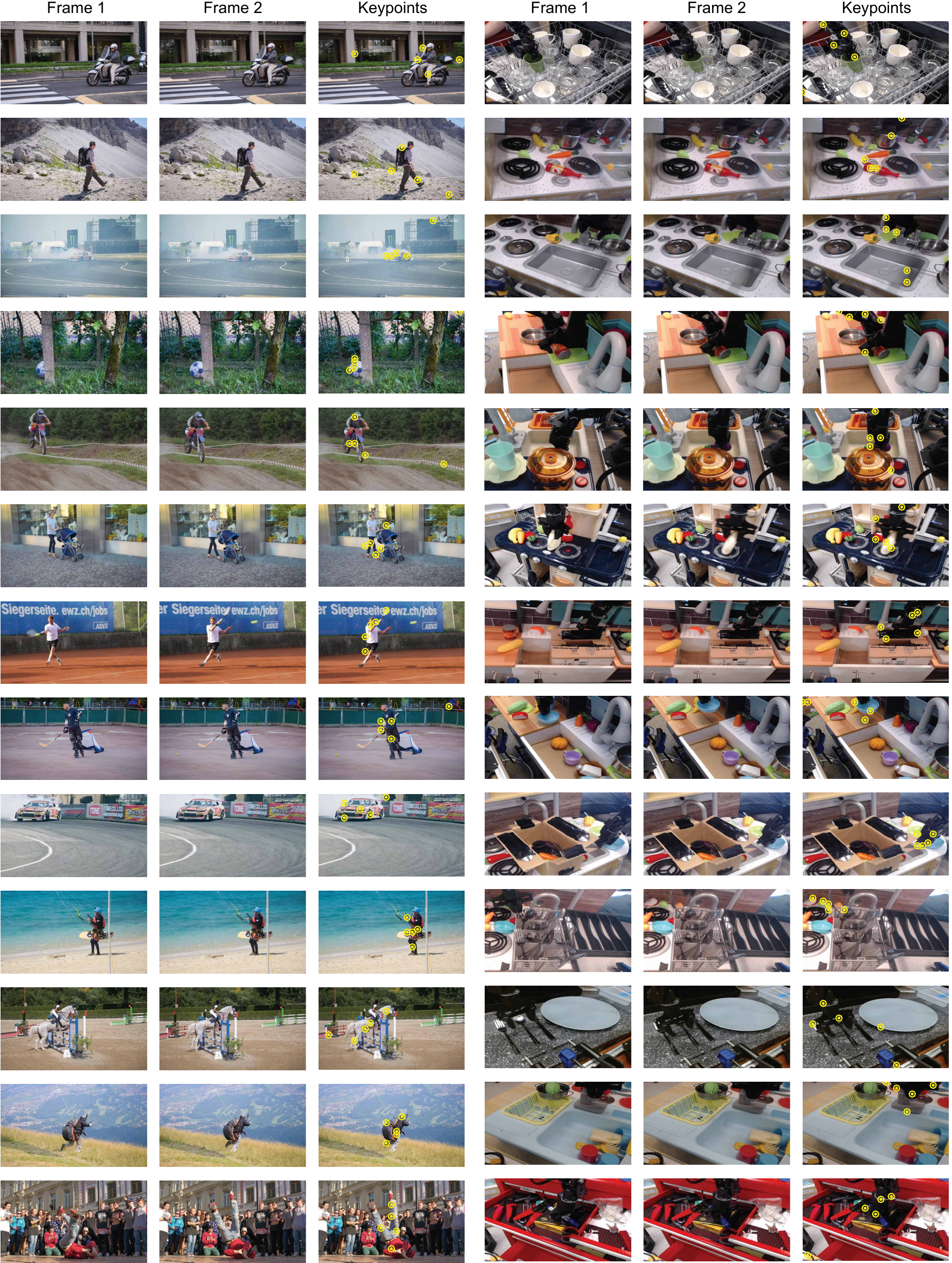}
  \caption{\textbf{Keypoints extracted by CWM.} We extract 5 keypoints for each example on DAVIS 2016 (left) and Brige dataset (right).~\pageref{figmention:keypoint}} 
  \label{fig:keypoints}
\end{figure*}

Figure \ref{fig:keypoints} shows additional qualitative results of the keypoints extracted on DAVIS 2016 \cite{Perazzi2016} and Bridge dataset \cite{ebert2021bridge}. We extract 5 keypoints for each example. Our procedure extracts dynamical RGB keypoints in the input frame pairs.

\subsection{Optical flow}






\label{supp_flow}

\subsubsection{Implementation details}
As originally proposed in \cite{bear2023unifying}, we simultaneously estimate optical flow at all locations in a frame pair $x_1, x_2 \!\in\! \mathbb{R}^{3\times H \times W}$ via the Jacobian of $\Psi$, denoted as $\mathcal{J}\Psi \!\in\! \mathbb{R}^{H \times W \times H \times W}$. The Jacobian of $\Psi$ assigns to element $(i, j, k, l)$ the predictor's change in output at location $(k,l)$ in the second frame due to an infinitessimal change at location $(i, j)$ in the first frame. Flow can be computed using $\mathcal{J}\Psi$ as the following:
    \begin{equation}
        \textbf{flow}(k, l) = \begin{cases}
                   \textbf{undefined } \text{(disocclusion)} & \quad \text{if } \textbf{argmax}_{i,j}|\mathcal{J}\Psi(i,j,k,l)| \ll 1 \\
                   (k,l) - \textbf{argmax}_{i,j}\mathcal{J}\Psi(i,j,k,l)  & \quad \text{otherwise}
              \end{cases}
    \end{equation}
with results averaged over several choices of visible patches in the masked second frame $x_2^\beta$. This is a tensorial operation that enables parallel computation for flow at all pixel locations, implemented practically using Jacobian-vector products available in Pytorch autograd. Note that this method detects disocclusion rather than occlusion, since no perturbation at any location in the first frame will cause a response at a point that becomes disoccluded in the second frame.

\subsubsection{Additional qualitative results} 
\begin{figure*}
  \centering
  \includegraphics[width=\linewidth]{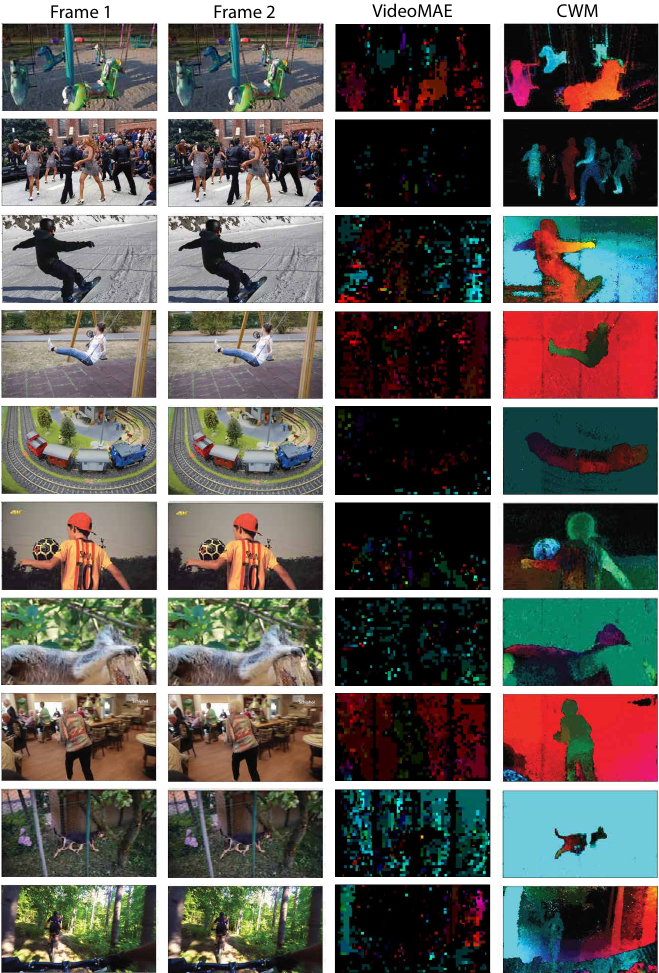}
  \vspace{0.05cm}
  \caption{\label{first}\textbf{Additional Optical Flows extracted on the DAVIS dataset}. We apply our flow extraction procedure to both CWM and VideoMAE predictors, and compare the extracted flows.~\pageref{figmention:flow}} 
  \label{fig:supp_DAVIS}
\end{figure*}

\begin{figure*}
  \centering
  \includegraphics[width=0.99\linewidth]{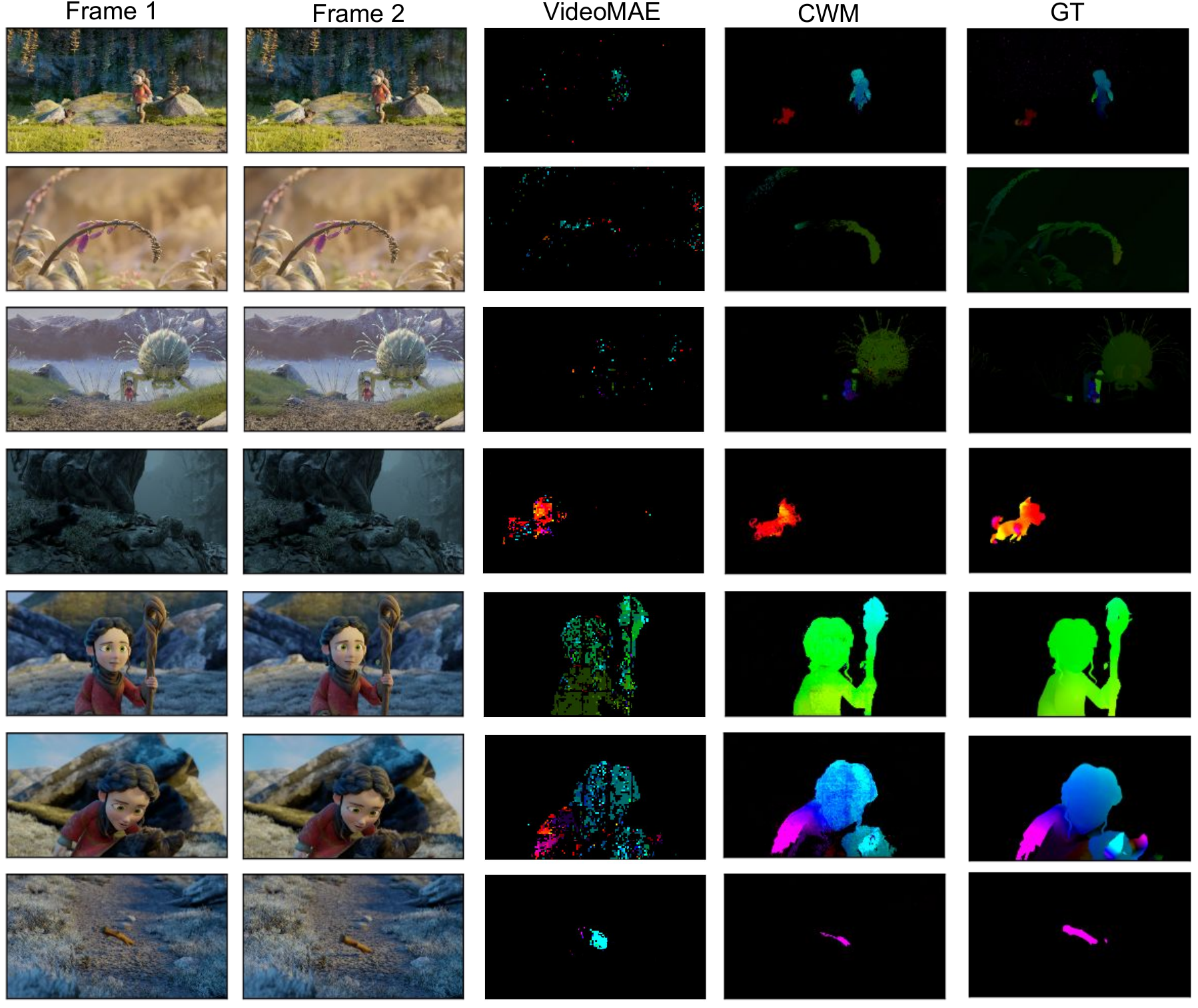}
  \caption{\label{first}\textbf{Additional Optical Flows extracted on the Spring dataset}.~\pageref{figmention:flow}}
  \label{fig:spring_results}
\end{figure*}

We show additional qualitative results on two distinct datasets: DAVIS 2016~\cite{davis2017} and a recent synthetic dataset SPRING~\cite{Mehl2023_Spring}. The results are shown in Figure \ref{fig:supp_DAVIS} and \ref{fig:spring_results} respectively. 
\label{figmention:flow}

\subsection{Segmentation}

\subsubsection{Single Spelke object extraction} To extract the Spelke object at a pixel location $(i,j)$ of a static image, we first create an intervention that simulates counterfactual motion by taking the patch at location $(i,j)$ and creating a new frame that is largely blank, but in which the content of the patch has been copied (e.g.
translated) to a new location $(i\!+\!\epsilon_1, j\!+\!\epsilon_2)$, where $\epsilon_1,\epsilon_2$ are location offsets randomly sampled within a radius $r\!>\!0$. We set $r$ to a fixed fraction (0.2) of the input image size. In addition, we can optionally create the appearance of stopping the counterfactual motion at a location $(i',j')$ by directly copying the patch at that location to the same location in the intervention without offset (or equivalently $r\!=\!0$). Adding the stop-motion patch allows the counterfactual query to isolate a single object, especially in a cluttered scene with multiple objects adjacent to and stacked on top of one another. In practice, different random choices of motion offset and stop-motion patches could potentially yield different counterfactual motion results. We sample 4 different interventions per pixel location, compute flow for the counterfactual motion, average the flow magnitudes of the different samples, and then threshold the mean flow magnitude map at 0.5 to obtain the binary segmentation map at pixel location $(i,j)$. We can use CWM to estimate optical flows, following procedures described in section \ref{supp_flow}. For faster extraction, we use RAFT \cite{teed2020raft} to estimate the optical flows of different samples.

We also find that the above procedure can be repeated iteratively to refine the segments further. Once a tentative segmentation map is obtained, we can sample more patches within the segment and add them to the intervention to simulate better counterfactual motion. At each iteration, we sample one patch within the segment and add it to the set of patches that simulates counterfactual motion. We additionally sample one patch outside the segment and add it to the set of stop-motion patches that simulates stopping the counterfactual motion. We set the number of iterations to 3 by default.

While the predictor $\Psi$ is trained with input resolution $224\!\times\!224$, Spelke object can be extracted at a different resolution by simply interpolating the position encoding correspondingly. We extract zero-shot segmentations on COCO training images at resolution $480\!\times\!480$ using the ViT-B/8 CWM model.

\vspace{-0.2cm}
\subsubsection{Multiple Spelke object extraction} To automatically discover multiple Spelke objects in a single image, we choose interventions at pixel locations based on a sampling probability distribution $\alpha$, which has high probability at pixel locations belonging to Spelke objects and low otherwise. Once a segment is discovered, we mask out the probability distribution values using the segments and repeat the process to discover the next object. 

We find two choices of sampling distribution work well. One is a movability distribution computed by sampling a few random interventions and averaging the motion responses across multiple predictions. The second choice is the prominence map computed by applying normalized cut to the patch-wise feature similarity matrix as proposed by CutLER \cite{wang2023cut}. In practice, the second approach yields slightly better qualitative segmentation results. Therefore, we choose the second approach as the default for computing the sampling probability distribution.


\subsubsection{Distillation} We follow the same procedure in the previous work CutLER \cite{wang2023cut} to distill segmentations extracted from a large task-agnostic pre-trained model into a smaller instance segmenter for faster and more robust segmentation. The extracted segmentations are used as pseudo annotations to train a downstream instance segmenter in a self-supervised manner.

\label{figmention:segment}
\subsubsection{Additional qualitative results} Figure \ref{fig:zero_shot_1} and \ref{fig:zero_shot_2} show more qualitative segmentation results of Spelke objects extracted by CWM on COCO training images, and compare them to those of other baseline methods FreeSOLO \cite{wang2022freesolo} and CutLER \cite{wang2023cut}. In each image, we set the maximum number of Spelke objects to be extracted as 3. Figure \ref{fig:maskrcnn_1} and \ref{fig:maskrcnn_2} show unsupervised segmentation results from the distilled instance segmenter. We show the results on COCO validation images.


\section{Dynamics understanding experiments}

\subsection{Physion benchmark}

\label{figmention:physion}
\subsubsection{Dataset details} 
\begin{figure*}
  \centering
  \begin{subfigure}[b]{0.95\linewidth}
    \includegraphics[width=\linewidth]{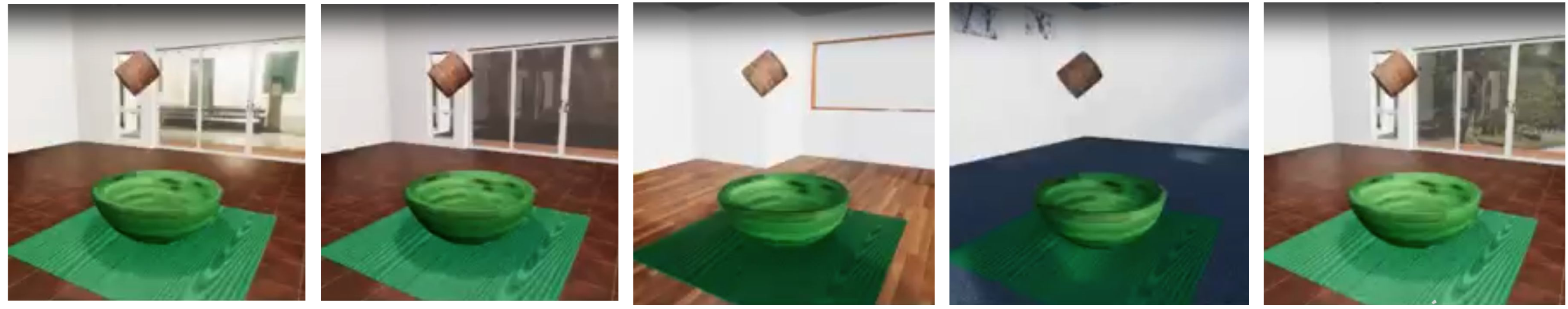} 
    \subcaption{Diverse lighting conditions}
    \label{fig:Lighting}
    \vspace{0.6cm}
  \end{subfigure}
  \begin{subfigure}[b]{0.44\linewidth}
    \includegraphics[width=\linewidth]{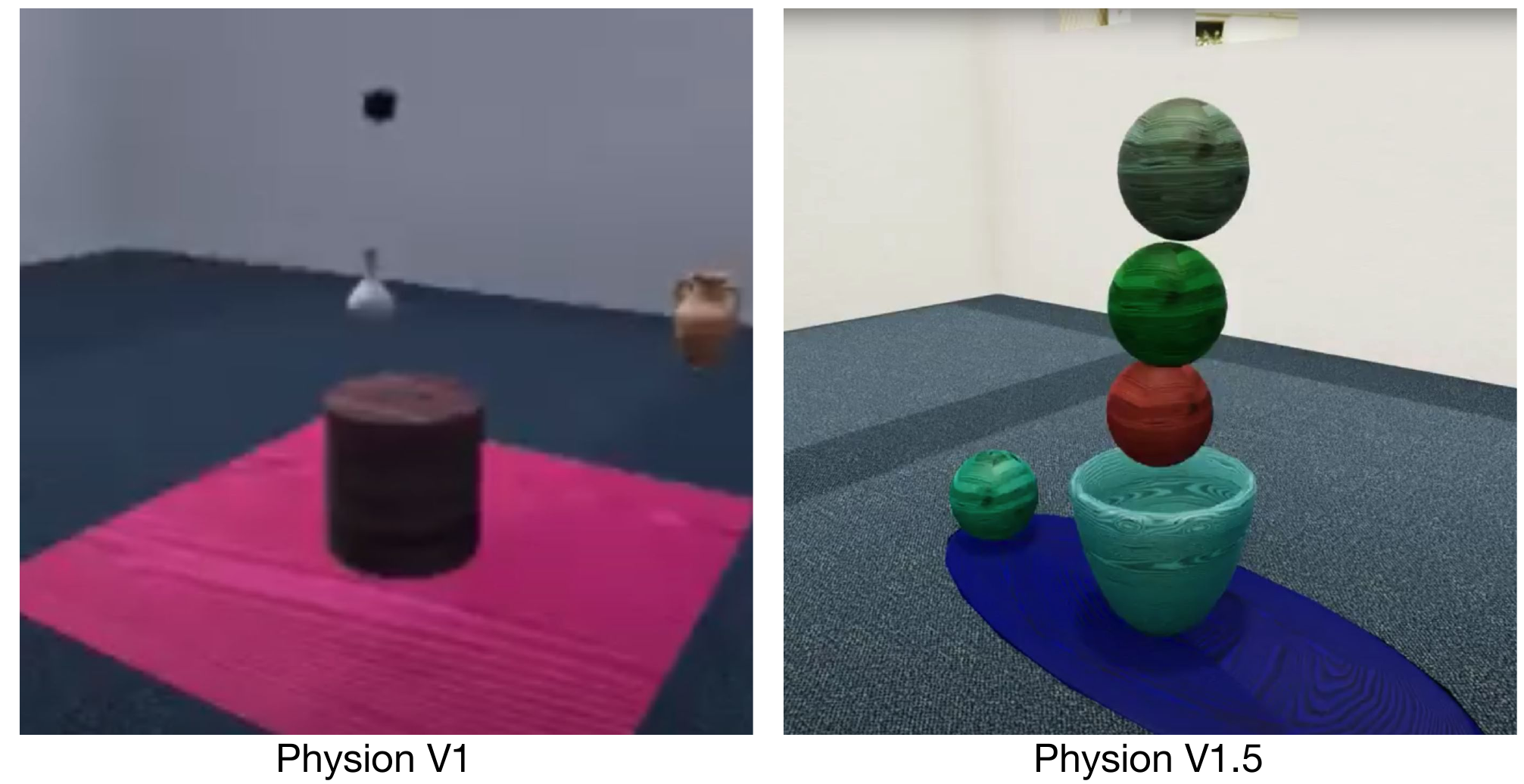} 
    \subcaption{Improved visual realism: high resolution rendering}
    \label{fig:Brightness}
    \vspace{0.6cm}
  \end{subfigure}\hspace{0.4cm}
  \begin{subfigure}[b]{0.45\linewidth}
    \includegraphics[width=\linewidth]{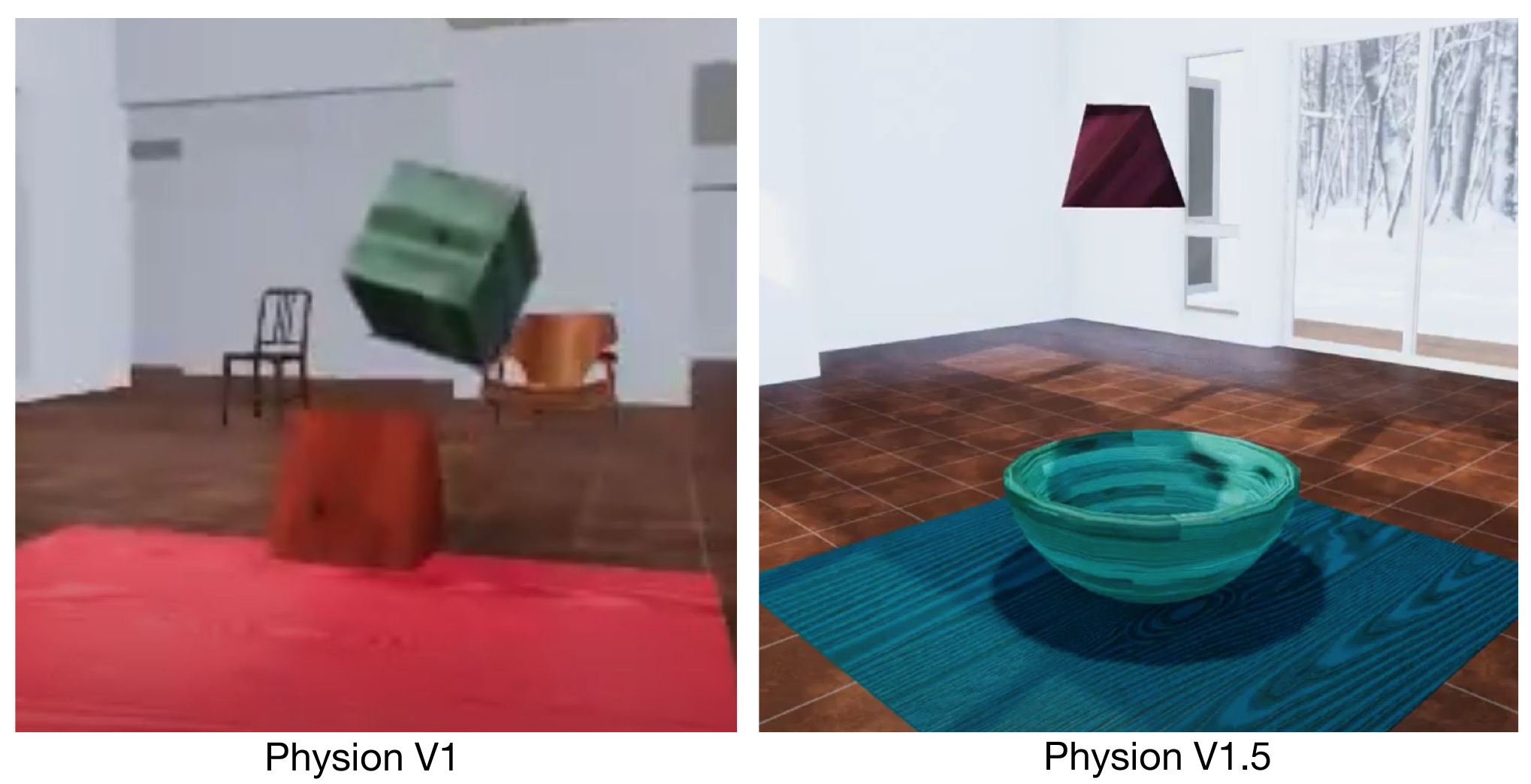} 
    \subcaption{Improved visual realism: rendering better shadows}
    \label{fig:Shadows}
    \vspace{0.6cm}
  \end{subfigure}
    \begin{subfigure}[b]{0.6\linewidth}
    \includegraphics[width=\linewidth]{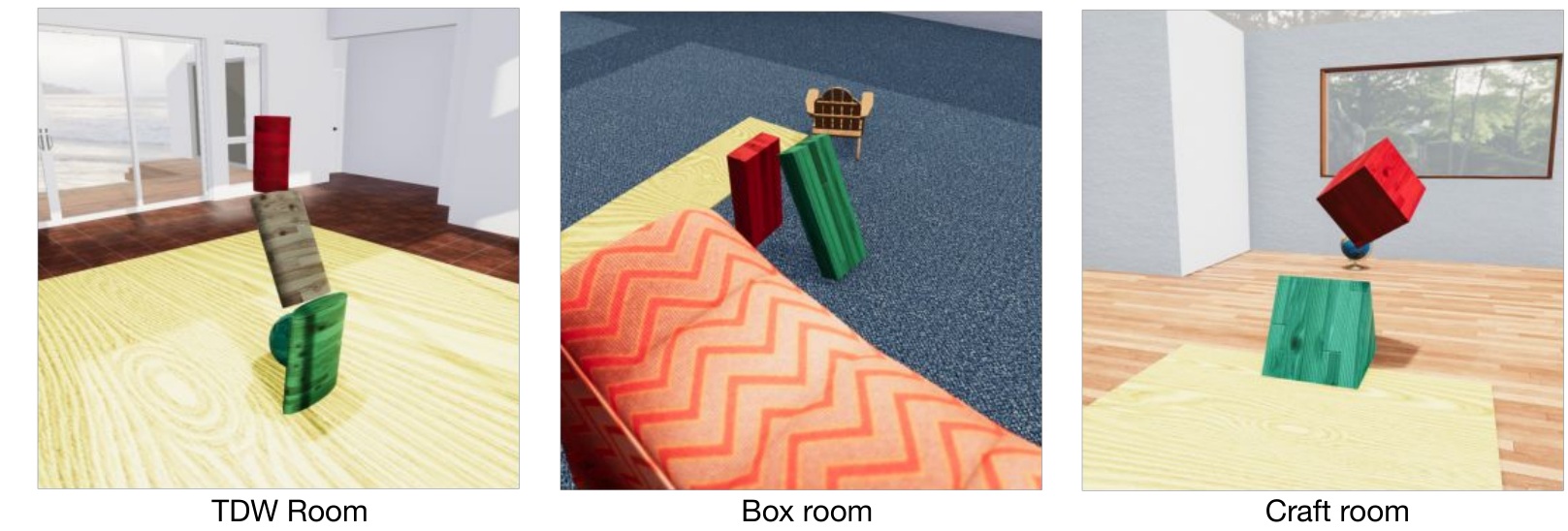}
    \subcaption{New indoor environments}
    \label{fig:Rooms}
  \end{subfigure}
  \caption{\textbf{Physion v1.5 features key improvements over v1} such as a) enhanced diversity of lighting conditions by employing HDRI skyboxes, b) higher resolution rendering, c) improved rendering of shadows and c) an additional indoor environment (``Craft room'').~\pageref{figmention:physion}}
  \label{fig:physion_improvements}
\end{figure*}

As discussed in the main text, we use the Physion v1.5 benchmark to evaluate CWM and baseline models on physical dynamics understanding. Physion v1.5 has several key improvements over Physion v1, which are illustrated in Figure \ref{fig:physion_improvements}. More specifically,  Physion v1.5 introduces another indoor environment, called the ``craft room'', in addition to the two environments featured in Physion v1 (see Figure~\ref{fig:Rooms}). Furthermore, v1.5 enhances the diversity of lighting conditions by employing a collection of 8 unique HDRI skyboxes, specifically designed to simulate various environmental lighting scenarios. This enhancement allows for dynamic time-of-day simulations in the room by adjusting the orientation of the skyboxes and directional lighting (see Figure~\ref{fig:Lighting}). Physion v1.5 also has improved rendering quality and photorealism in comparison to v1 (see Figure~\ref{fig:Brightness} and~\ref{fig:Shadows}). The physics simulations and rendering are done using the ThreeDWorld simulation platform~\cite{gan2020threedworld}. 

Physion v1.5 comprises seven distinct physical scenarios, including collide, drop, dominoes, contain, roll, support, and link. This version comprehensively demonstrates various aspects and challenges of rigid body physics (see Figure~\ref{fig:physion_ocp_results_1} for examples of each scenario). We train and test the linear classifier model (as outlined in Section~\textcolor{red}{4.1} of the main text) on all seven scenarios. 

\vspace{-0.2cm}

\subsubsection{Per-scenario OCP and OCD results.} In the main text, we presented the OCP and OCD scores averaged across all seven scenarios. Now, we provide a detailed breakdown of performance for each specific scenario, as shown in Table~\ref{tab:main_results_ocp_sce} and Table~\ref{tab:main_results_ocd_sce}. 

\vspace{-0.2cm}
\subsubsection{Qualitative comparisons on OCP and OCD task.} 
\begin{figure*}
  \centering
  \includegraphics[width=0.98\linewidth]{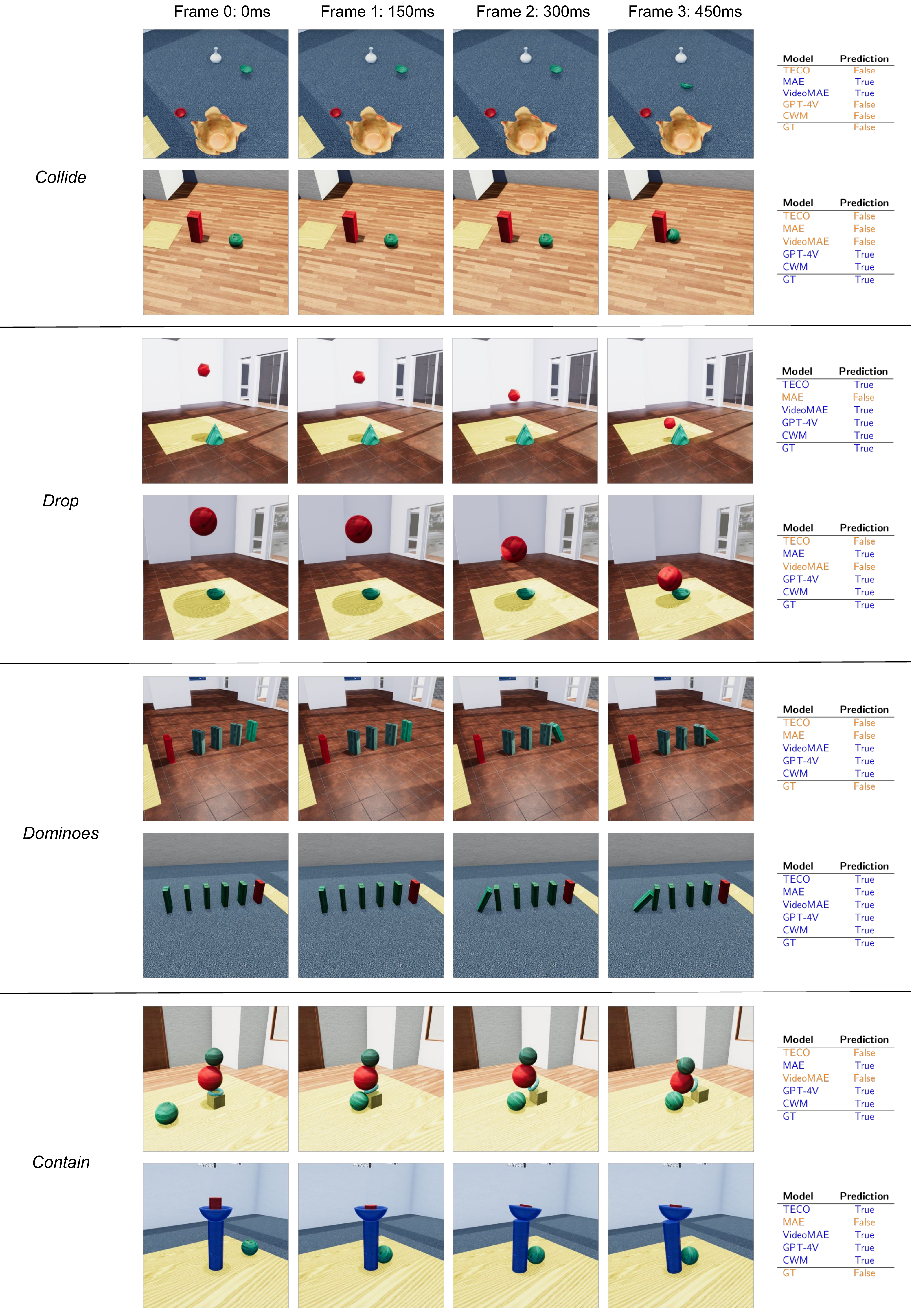}
  \caption{\textbf{Model predictions on the OCP task.} The input frames sampled at a frame gap of 150ms are shown on the left and the model predictions are shown on the right. We compare against the best performing models in each model category outlined in Table~\ref{tab:main_results_ocp_sce}. ~\pageref{figmention:physion}} 
  \label{fig:physion_ocp_results_1}
\end{figure*}

\begin{figure*}
  \centering
  \includegraphics[width=0.98\linewidth]{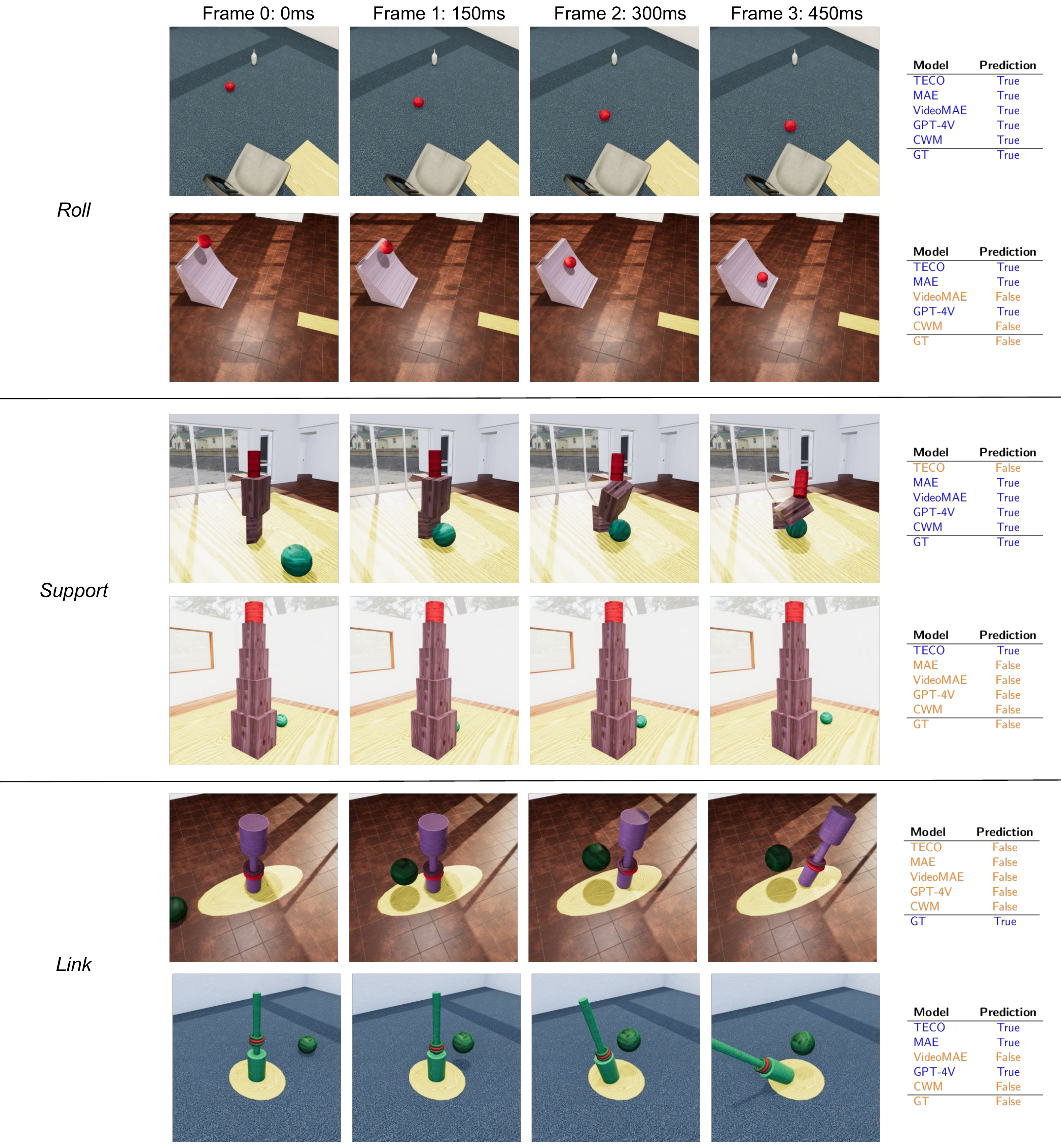}
  \caption{\textbf{More model predictions on the OCP task}. ~\pageref{figmention:physion}} 
  \label{fig:physion_ocp_results_2}
\end{figure*}

\begin{figure*}
  \centering
  \includegraphics[width=0.98\linewidth]{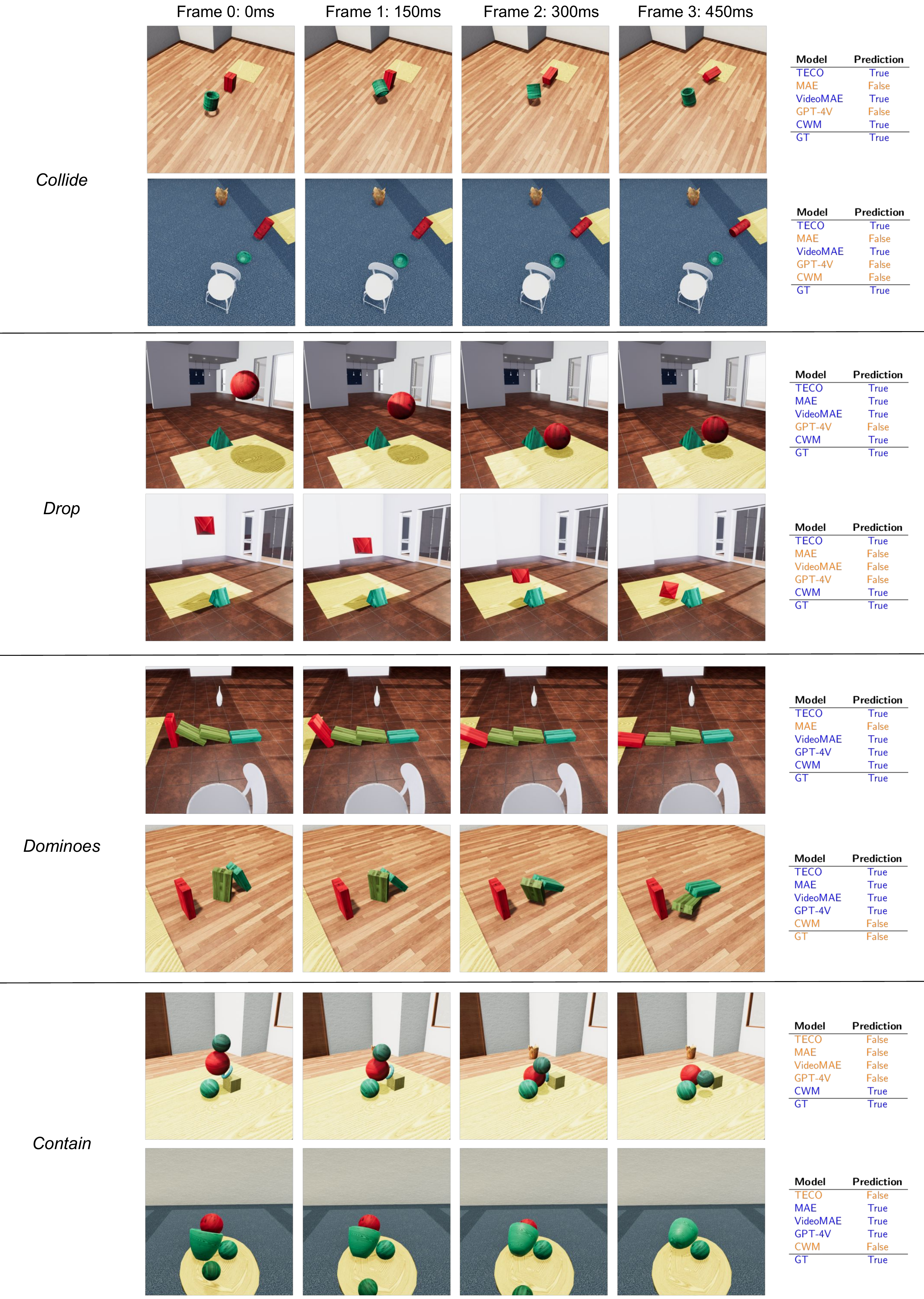}
  \caption{\textbf{Model predictions on the OCD task.} The input frames sampled at a frame gap of 150ms are shown on the left and the model predictions are shown on the right. We compare against the best performing models in each model category outlined in Table~\ref{tab:main_results_ocd_sce}.~\pageref{figmention:physion}} 
  \label{fig:physion_ocd_results_1}
\end{figure*}

\begin{figure*}
  \centering
  \includegraphics[width=0.98\linewidth]{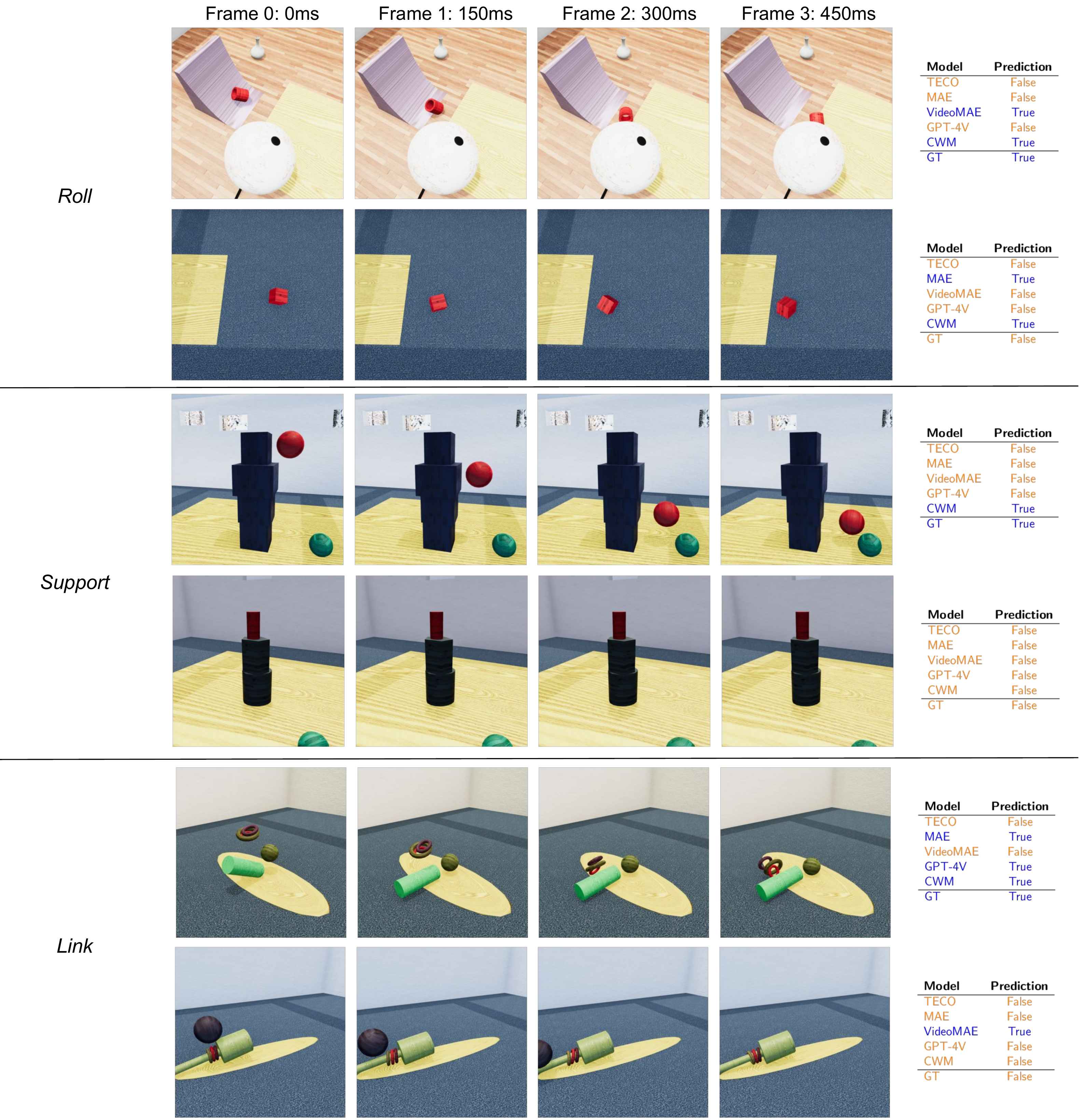}
  \caption{\textbf{More model predictions on the OCD task}. ~\pageref{figmention:physion}} 
  \label{fig:physion_ocd_results_2}
\end{figure*}

We supplement our quantitative results on the Physion v1.5 tasks with qualitative visualizations in Figures~\ref{fig:physion_ocp_results_1},~\ref{fig:physion_ocp_results_2},~\ref{fig:physion_ocd_results_1} and~\ref{fig:physion_ocd_results_2}. We show several example inputs for each scenario, along with the classification results of a linear probe on top of the CWM model, compared with those of leading baselines in each model category outlined in Table~\textcolor{red}{1} in the main text.  

\vspace{-0.2cm}
\subsubsection{Integrating vision structures for OCP and OCD tasks.} In the main text, we demonstrate how zero-shot vision structures extracted from CWM improve OCP and OCD performance on Physion v1.5. This section describes the details of how these structures are integrated prior to linear probing on downstream tasks. Keypoint information is integrated by incorporating patch features at the keypoint locations. Optical flow information is integrated by providing $8\times8$ patches of optical flow value at the keypoint locations. Finally, segment information is integrated by using the segments to pool the feature map and incorporate the aggregated features for linear probing.

The integration process for segments is detailed as follows: Let $F \!\in \!\mathbb{R}^{H\times W \times D}$ be the feature map of an input frame from the last layer of the ViT encoder, where \( H \), \( W \), and \( D \) represent height, width, and channel dimension. Suppose we have binary segmentation masks \( S \) with dimension \( N \times H \times W \), where \( N \) denotes the number of masks. We compute the set of aggregated feature vectors for each of the segments,



\begin{equation}
    F_{agg} = \left\{ \frac{\sum_{i,j} S^n_{ij}F_{ij}}{\sum_{i,j} S^n_{ij}} \,\mid \, 1 \le n \le N \right\}
\end{equation}

where \( S^n \) is the \( n \)-th segmentation mask and $F_{ij}$ is the feature vector at location $(i,j)$. These aggregated feature vectors, along with the keypoint patch features and flow patches, are concatenated with the original feature map $F$ before being fed into the linear classifier. 








\subsection{GPT4-Vision prompting}

\label{figmention:gpt4v}
\subsubsection{GPT-4V prompting methodology and results.} 
\begin{figure*}
  \centering
  \begin{subfigure}[b]{0.99\linewidth}
    \includegraphics[width=\linewidth]{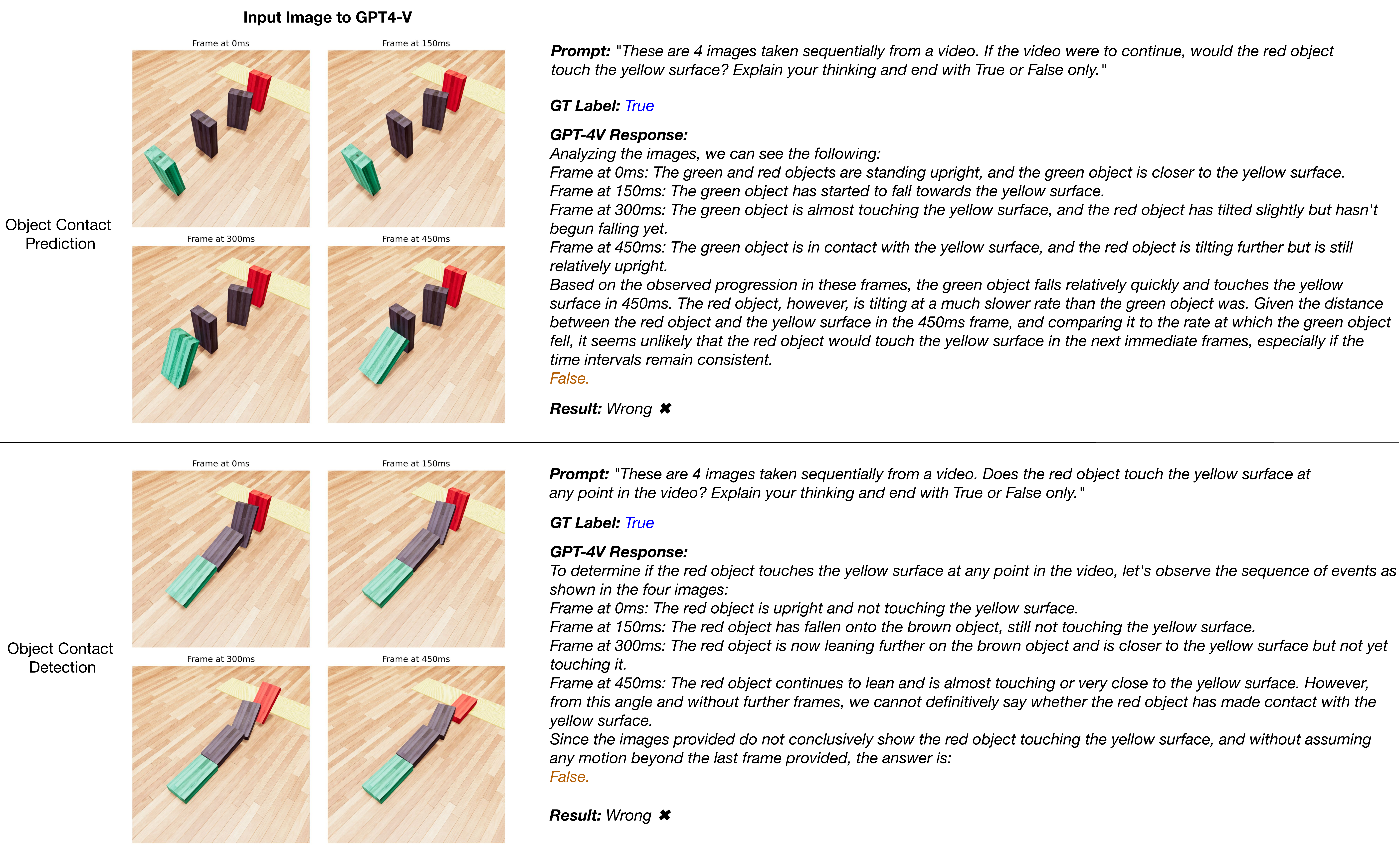}
    \subcaption{Querying method: RGB frames only}
    \label{fig:sub1}
  \end{subfigure}
  \begin{subfigure}[b]{0.99\linewidth}
    \includegraphics[width=\linewidth]{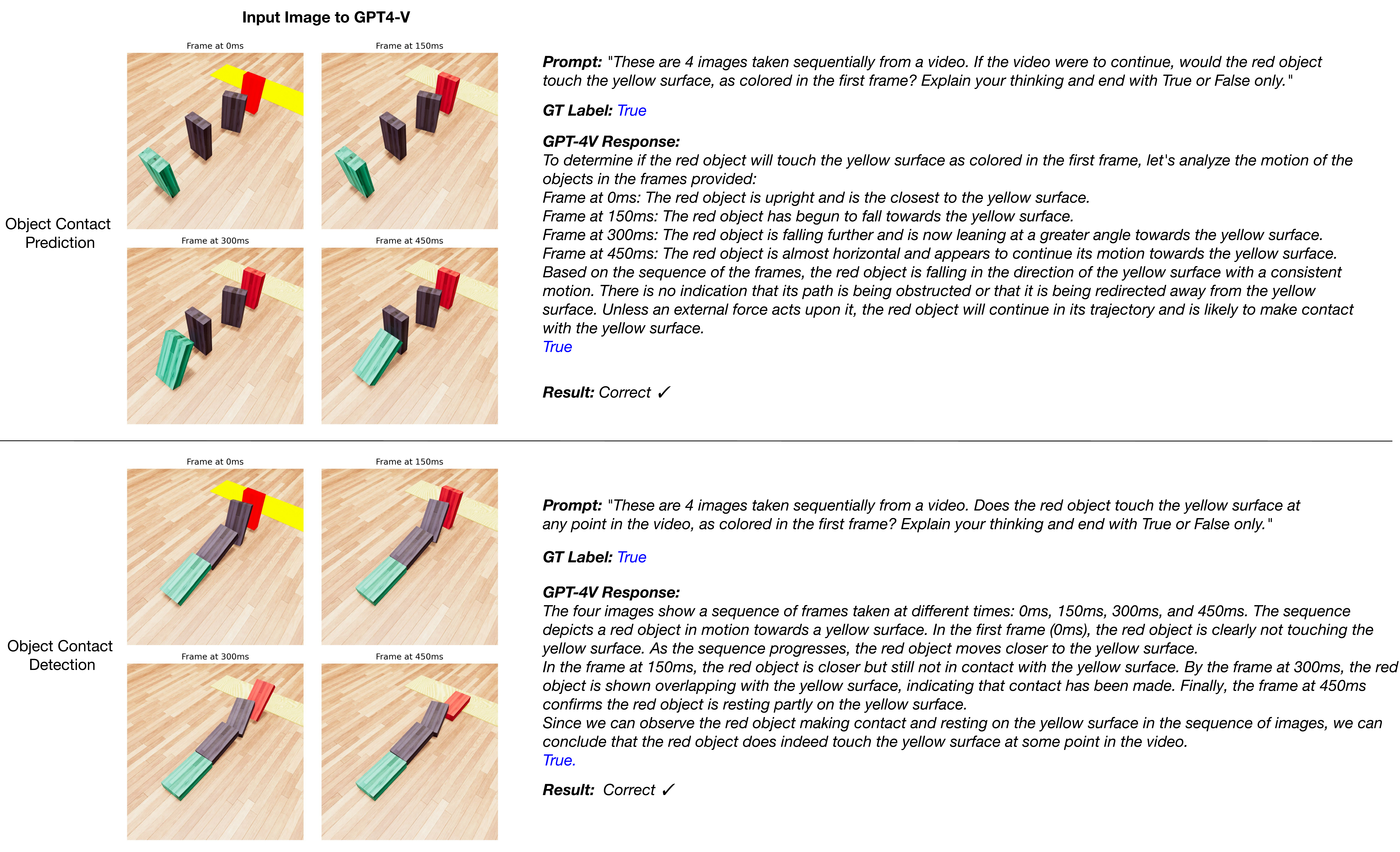} 
    \subcaption{Querying method: RGB frames with objects of interest highlighted}
    \label{fig:sub2}
  \end{subfigure}
  \caption{\textbf{More GPT-4V results}.~\pageref{figmention:gpt4v}}
  \label{fig:gpt_results_2}
\end{figure*}

\begin{figure*}
  \centering
  \begin{subfigure}[b]{0.99\linewidth}
    \includegraphics[width=\linewidth]{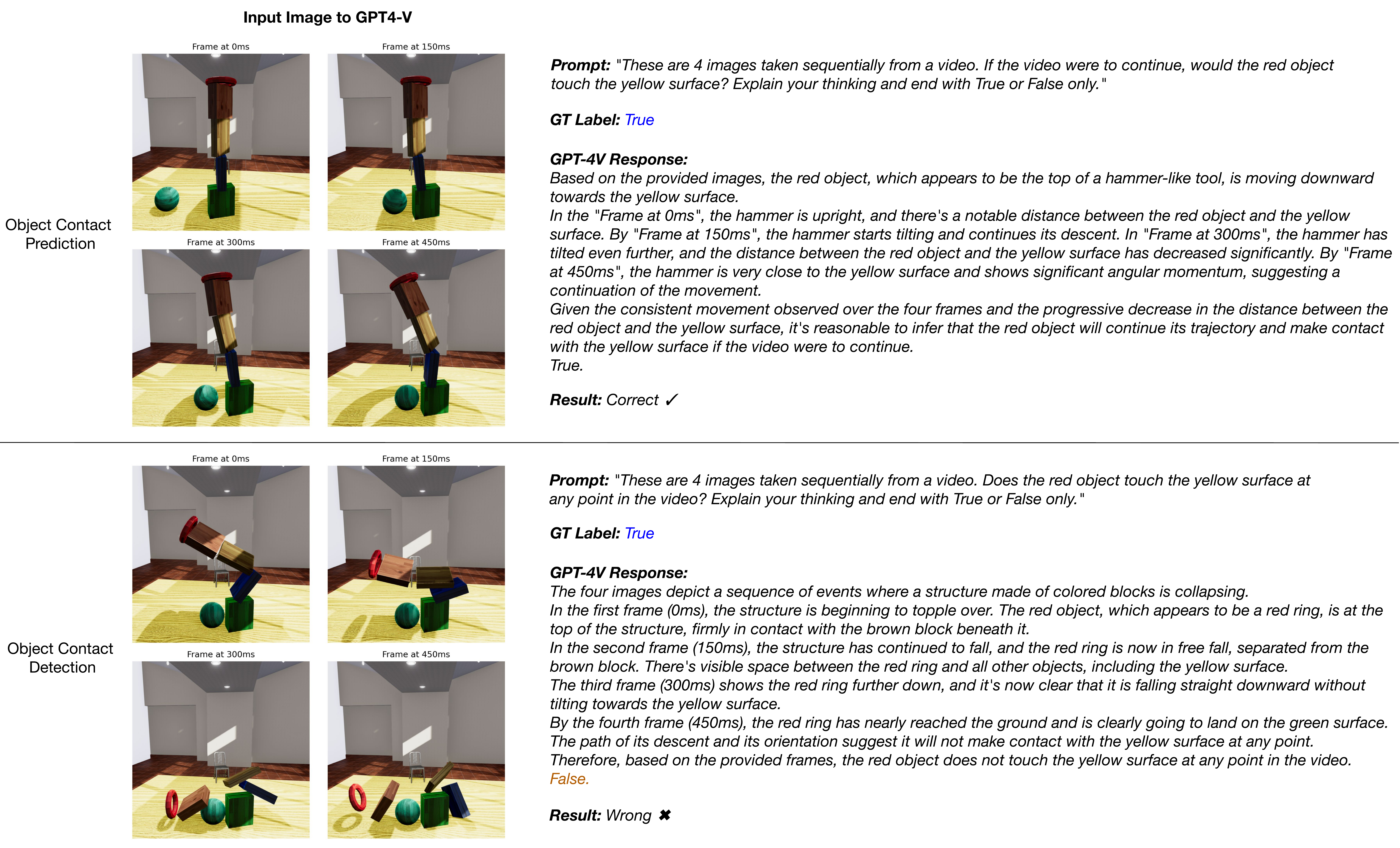}
    \subcaption{Querying method: RGB frames only}
    \label{fig:sub1}
  \end{subfigure}
  \begin{subfigure}[b]{0.99\linewidth}
    \includegraphics[width=\linewidth]{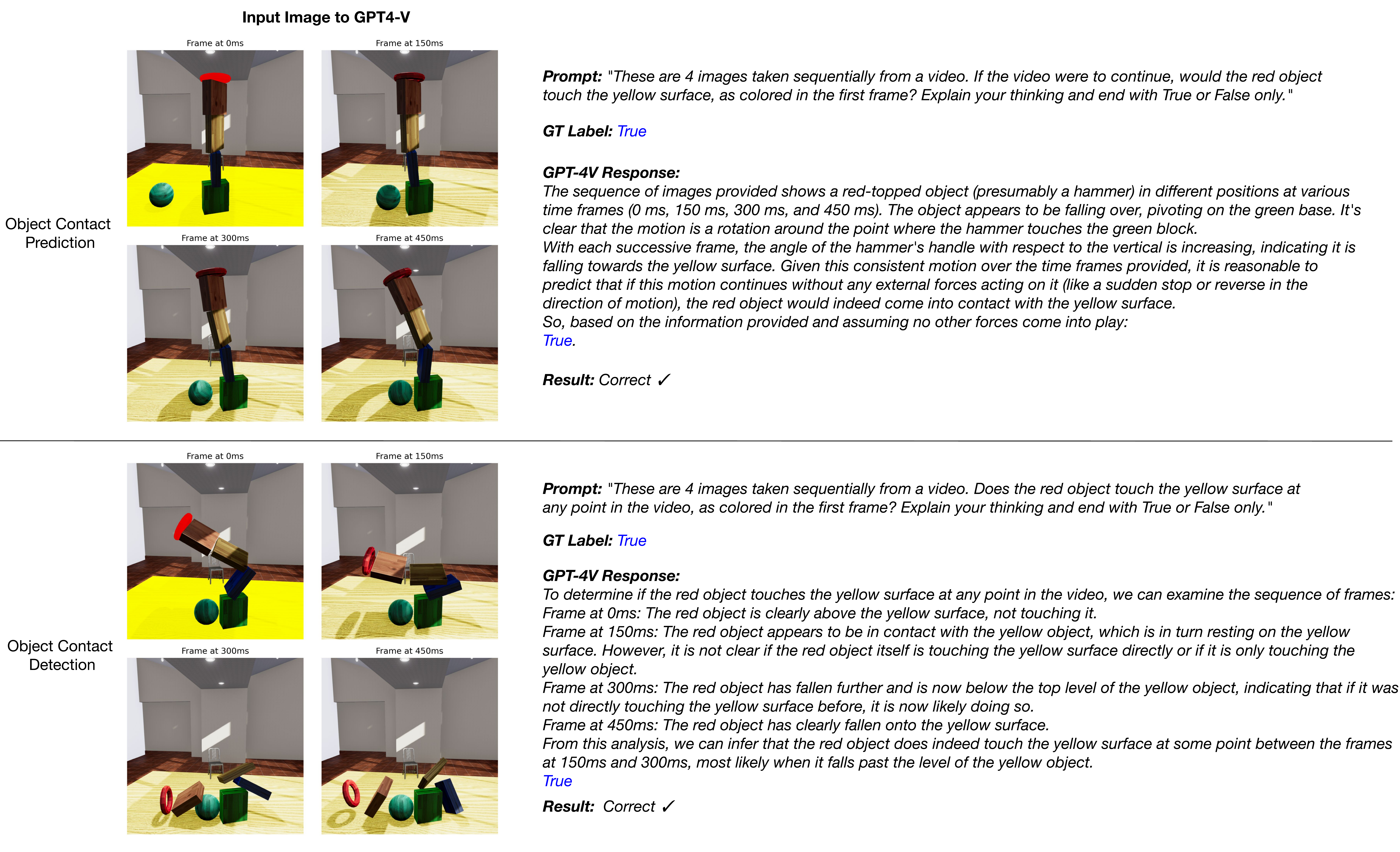} 
    \subcaption{Querying method: RGB frames with objects of interest highlighted}
    \label{fig:sub2}
  \end{subfigure}
  \caption{\textbf{More GPT-4V results}.~\pageref{figmention:gpt4v}}
  \label{fig:gpt_results_3}
\end{figure*}

\begin{figure*}
  \centering
  \begin{subfigure}[b]{0.99\linewidth}
    \includegraphics[width=\linewidth]{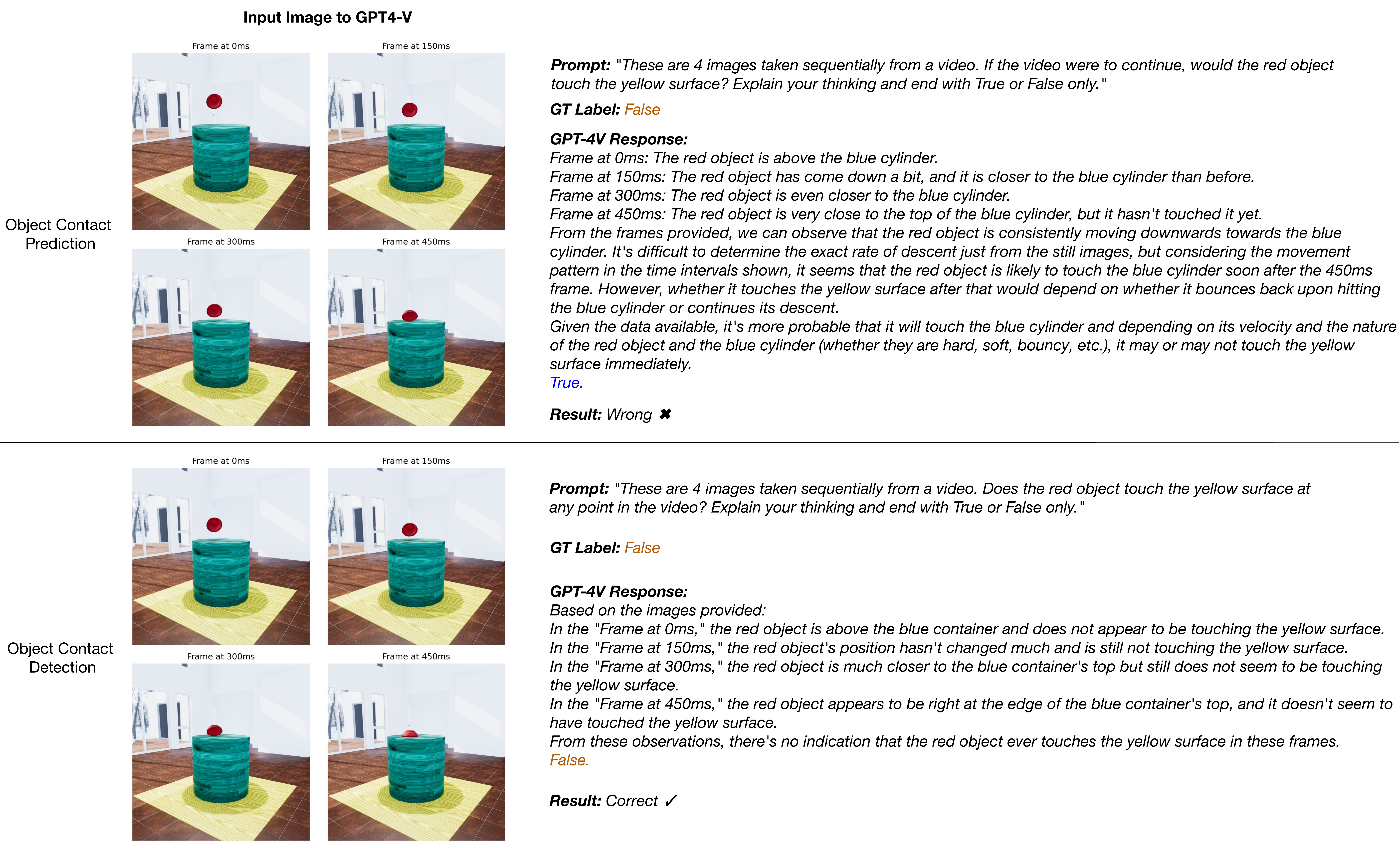}
    \subcaption{Querying method: RGB frames only}
    \label{fig:sub1}
  \end{subfigure}
  \begin{subfigure}[b]{0.99\linewidth}
    \includegraphics[width=\linewidth]{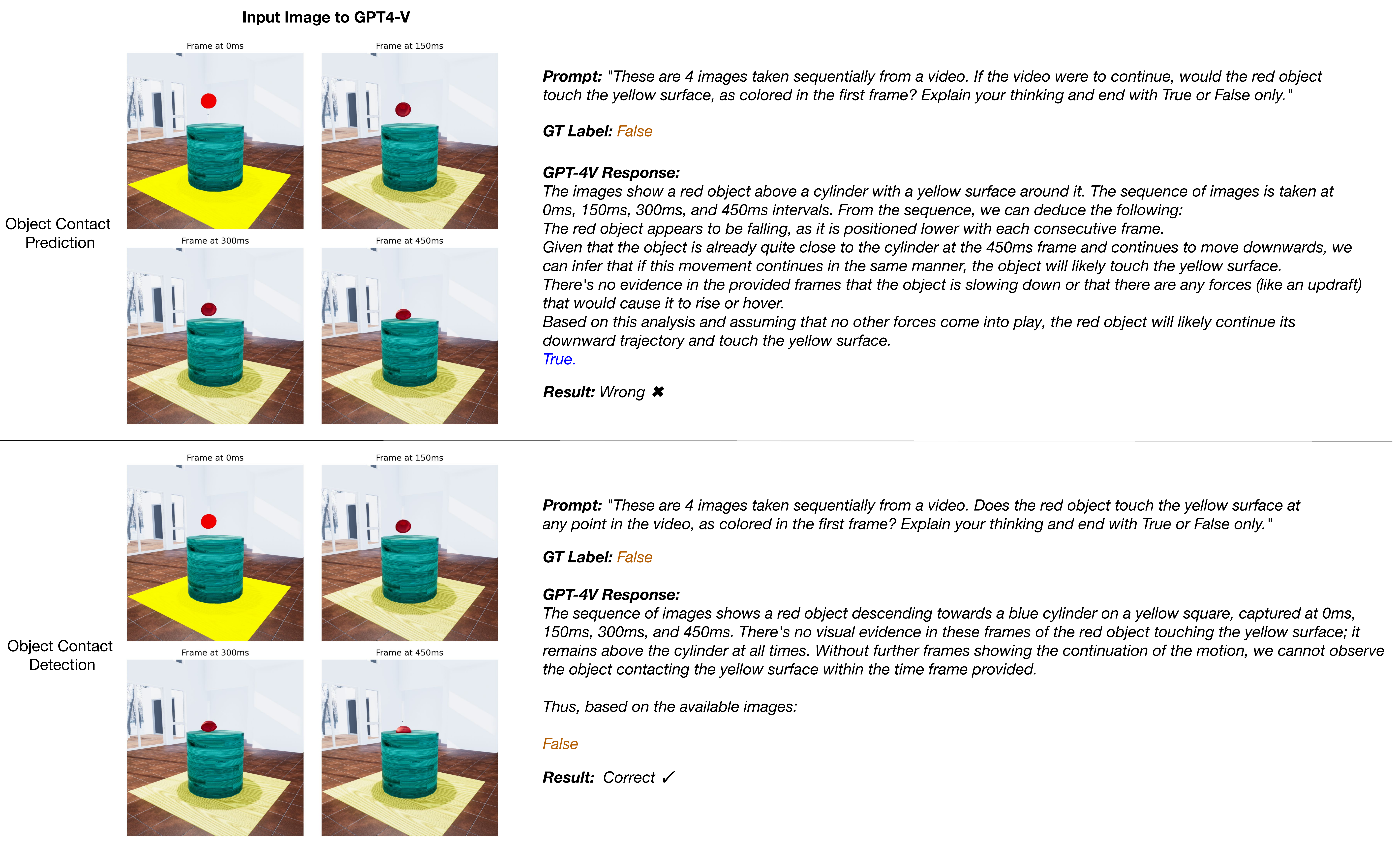} 
    \subcaption{Querying method: RGB frames with objects of interest highlighted}
    \label{fig:sub2}
  \end{subfigure}
  \caption{\textbf{More GPT-4V results.}~\pageref{figmention:gpt4v}}
  \label{fig:gpt_results_4}
\end{figure*}

In Figures~\ref{fig:gpt_results_1},~\ref{fig:gpt_results_2},~\ref{fig:gpt_results_3} and ~\ref{fig:gpt_results_4} we report a few results from testing GPT-4V on the OCP and OCD tasks in Physion v1.5. In each figure, the prompt image is shown on the left and the prompt text along with the GT label and response is shown on the right. To construct the image prompt, we tile four successive video frames (sampled at a frame gap of 150ms) into a $2\!\times\!2$ image with four panels. Each panel contains an RGB frame titled with its timestamp. Following the methodology used for evaluating vision-only models, the objects of interest for the contact-related queries are rendered with red and yellow textures to provide visual cues for the model. For OCP, the text prompt used was “These are 4 images taken sequentially from a video. If the video were to continue, would the red object touch the yellow surface? Explain your thinking and end with True or False only”. For OCD, the prompt used was “These are 4 images taken sequentially from a video. Does the red object touch the yellow surface at any point in the video? Explain your thinking and end with True or False only”. GPT4-V achieves 52.9\% accuracy for OCP and 54.7\% accuracy for OCD. For OCP, the model predicted “contact” 78.7\% of the time, while for OCD the model predicted “no contact” 81.5\% of the time. 
\begin{table}[t]
  \centering
  \footnotesize
    \caption{\textbf{GPT-4V performance on the Physion v1.5 tasks using two different promoting strategies}: a) with RGB frames only and b) RGB frames with ground truth segment map overlayed on the objects of interest.}
  \begin{tabular}{lcc}
    \toprule
    Prompting Method & OCP $\uparrow$ & OCD $\uparrow$ \\
    \midrule
    RGB frames  & 52.9 & 54.7 \\
    RGB frames + GT segment overlay & \textbf{58.3} & \textbf{67.5} \\
    \bottomrule
  \end{tabular}

  \label{tab:gpt4v_promoting} 
\end{table}

\vspace{-0.2cm}
\subsubsection{Alternate querying methods.} 
\begin{figure*}
  \centering
  \begin{subfigure}[b]{0.97\linewidth}
    \includegraphics[width=\linewidth]{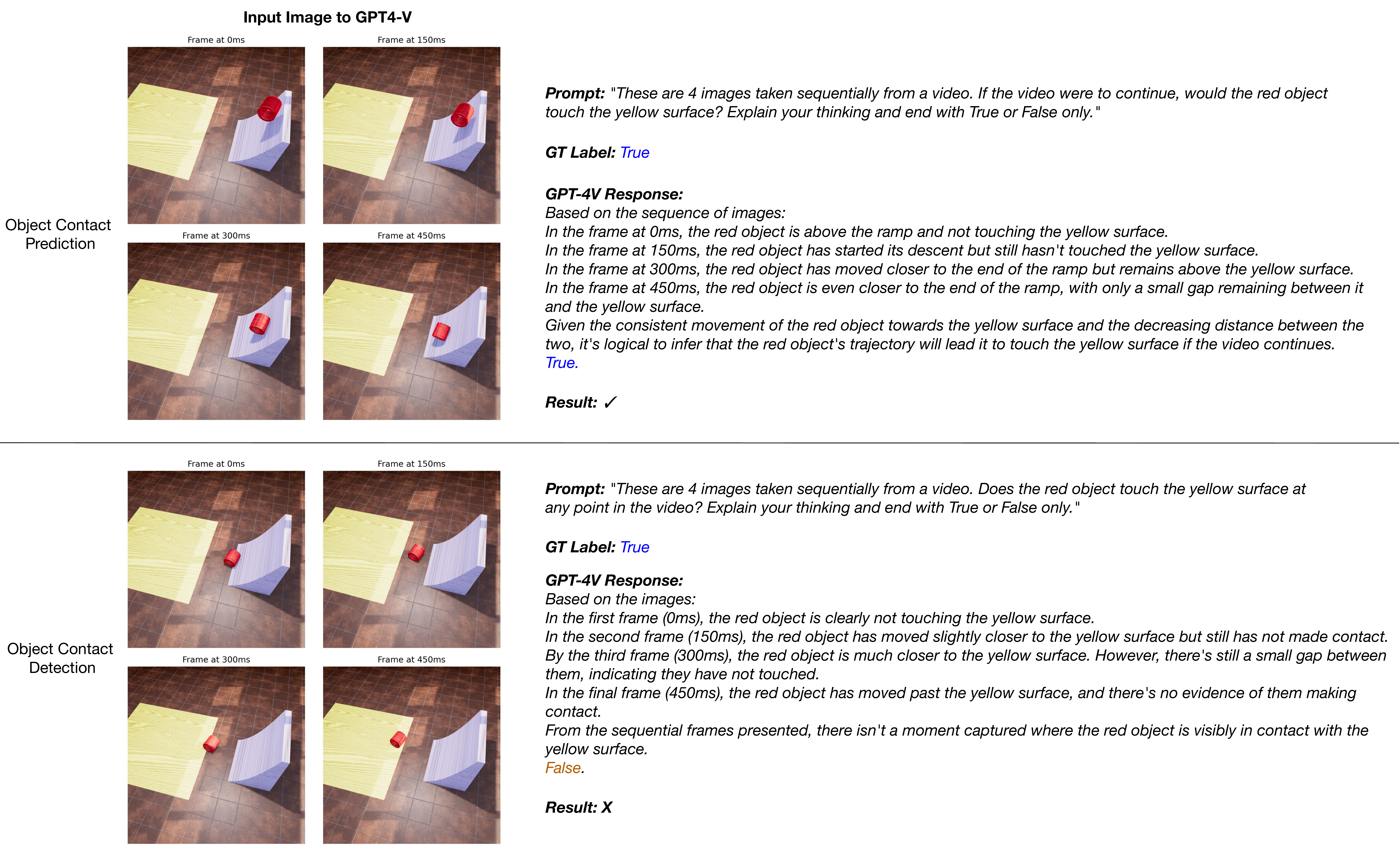}
    \subcaption{Querying method: RGB frames only}
    \label{fig:sub1}
  \end{subfigure}
  \begin{subfigure}[b]{0.97\linewidth}
    \includegraphics[width=\linewidth]{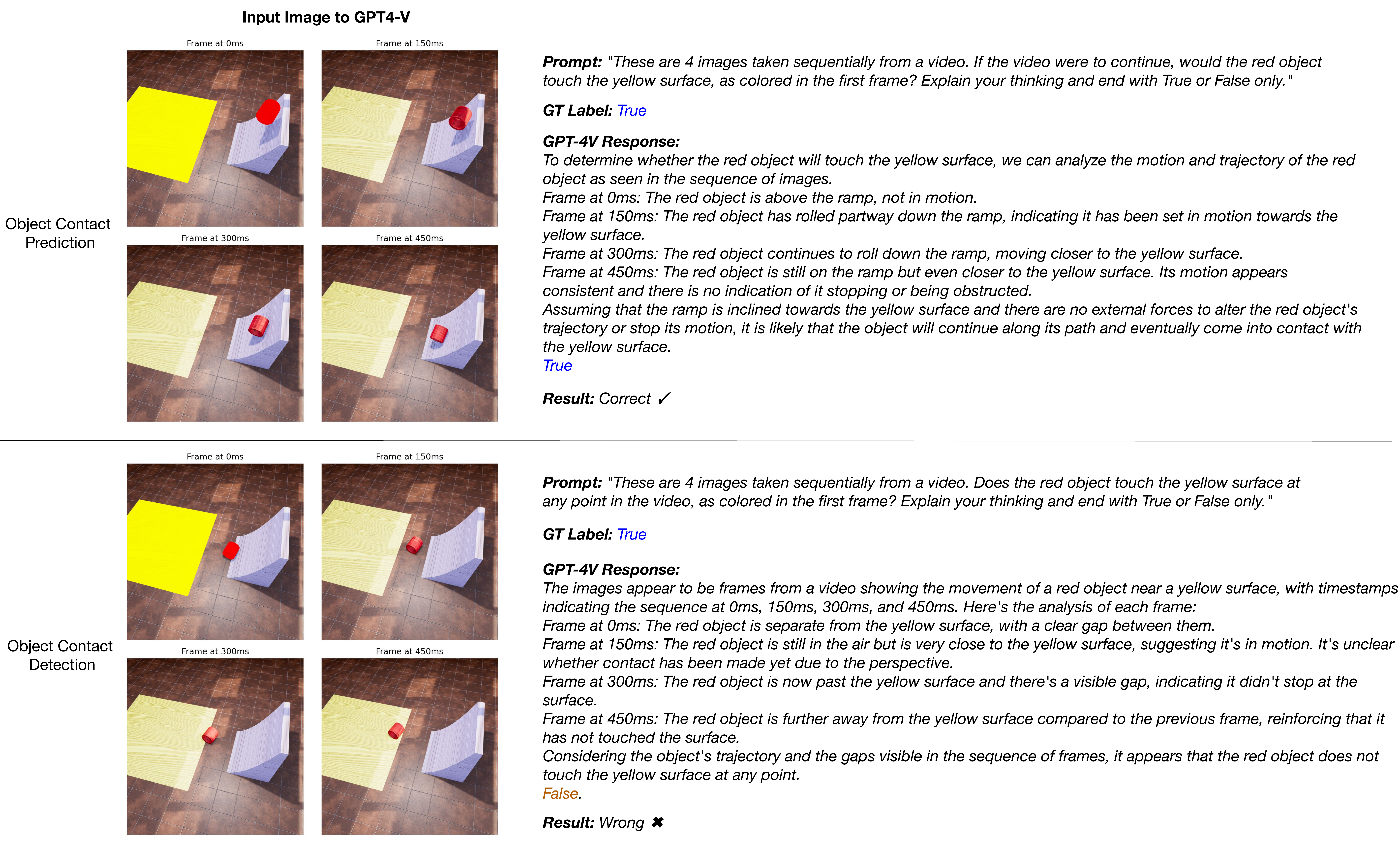} 
    \subcaption{Querying method: RGB frames with objects of interest highlighted}
    \label{fig:sub2}
  \end{subfigure}
  \caption{\textbf{GPT-4V results on the OCP and OCD tasks in Physion v1.5.} The input image prompt is shown on the left and the text prompt, GPT-4V response and results are shown on the right. We show the results of two different querying strategies: a) with RGB frames only and b) RGB frames with ground truth segment map overlayed on the objects of interest.~\pageref{figmention:gpt4v}}
  \label{fig:gpt_results_1}
\end{figure*}

Additionally, we experiment with an alternative querying method to explore the limit of GPT4-V in dynamics understanding. In addition to rendering the objects of interest in red and yellow texture, we apply a bright red and yellow ground-truth segmentation overlay on them to focus the model's attention on these objects (see Figure~\ref{fig:gpt_results_1} for visualizations). As shown in Table~\ref{tab:gpt4v_promoting}, we find that this querying strategy improves the OCP from 52.9\% to 58.3\% and the OCD from 54.7\% to 67.5\%. In the main text, we have also demonstrated a parallel phenomenon with CWM, where integrating segment information led to enhanced performance in related tasks (Table~\textcolor{red}{3} in main text). These observations highlight the importance of vision structures such as segmentation in downstream tasks associated with physical dynamics understanding. 

\vspace{-0.2cm}
\subsubsection{Implementation details.} We employ four GPT-4V accounts and retrieve the results using Selenium~\cite{selenium}, a browser automation tool. Our methodology adheres to the code framework available at\footnote{\hyperlink{https://github.com/Michelangelo27/chatgpt\_selenium\_automation}{https://github.com/Michelangelo27/chatgpt\_selenium\_automation}}.  
\section{Evaluating CWM on additional benchmarks}

\begin{table}[t]
    \centering
    \footnotesize
    \caption{Evaluation on additional benchmarks. (a) Activity recognition on Something-Something V2, (b) IntPhys intuitive physics benchmark.}
    \begin{minipage}{0.5\textwidth}
        \centering
        \begin{tabular}{lcc}
            \toprule
            Model & frames & accuracy $\uparrow$ \\
            \midrule
            VideoMAE & 16 & \textbf{54.7} \\
            VideoMAE* & 3 & 51.3 \\
            \midrule
            CWM & 3 & 54.2 \\
            \bottomrule
        \end{tabular}
        \subcaption{\small Activity recognition on SSv2}
        \label{tab:ssv2}
    \end{minipage}
    \hfill
    \begin{minipage}{0.48\textwidth}
        \centering
        \begin{tabular}{lccc}
            \toprule
            Method & B1$\downarrow$ & B2$\downarrow$ & B3$\downarrow$ \\
            \midrule
            VideoMAE & 0.40 & 0.23 & 0.30 \\
            VideoMAE\textsuperscript{*} & 0.46 & 0.30 & 0.30 \\
            VideoMAEv2 & 0.36 & 0.30 & 0.36 \\
            \midrule
            CWM & \textbf{0.36} & \textbf{0.20} & \textbf{0.26} \\
            \bottomrule
        \end{tabular}
        \subcaption{\small IntPhys}
        \label{table:intphys}
    \end{minipage}

\end{table}

\subsection{Activity Recognition}
We evaluate CWM on the Something-Something V2 benchmark for activity recognition and report the results in Table \ref{tab:ssv2}. To obtain model predictions we train an attentive probe on the feature representation similar to the setup used in V-JEPA~\cite{bardes2024revisiting}. We find that CWM outperforms VideoMAE trained with the same number of input frames (i.e VideoMAE*) by a fair margin and is comparable to the standard VideoMAE trained on a longer context window of 16 frames. 

\subsection{IntPhys}

In the main text we show evaluations on Physion as it is by far the most challenging and comprehensive benchmark in the literature, consisting of realistic 3D simulations from diverse physical scenarios. Here, we evaluate on IntPhys \cite{riochet2021intphys} which is a complementary benchmark to Physion with photorealistic simulations of various intuitive physics tasks. Table \ref{table:intphys} reports the relative error metric in IntPhys \cite{riochet2021intphys} evaluation on the validation split. For all models, we use the future frame reconstruction error to obtain a plausibility score for a given video. The evaluation comprises of three tasks: B1 tests for object permanence, B2 tests for shape constancy, and B3 tests for spatio-temporal continuity. We find that CWM, despite trained with less number of frames, outperforms VideoMAE. CWM (86M) outperforms VideoMAEv2 (1.1B) despite having fewer parameters. For a fair comparison, we further evaluate VideoMAE\textsuperscript{*} which is trained on the same number of frames as CWM but with tube masking instead of temporally factored masking. The performance gain from CWM further validates the benefit of the temporally-factored masking policy.

\clearpage 

\begin{subfigures}

\begin{figure*}
  \centering
  \includegraphics[width=\linewidth]{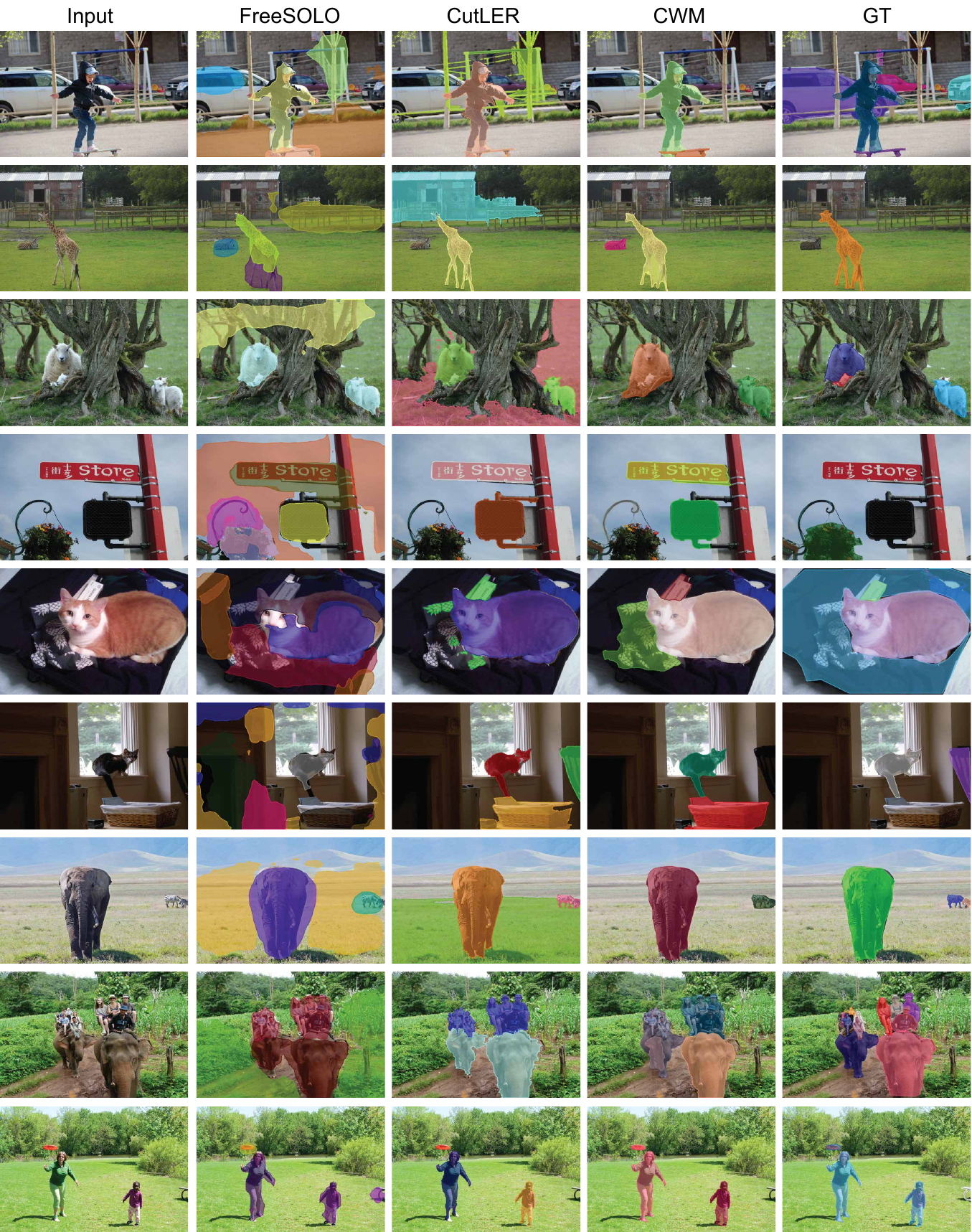}
  \caption{\label{first}\textbf{Pseudo-masks extracted in a zero-shot manner on COCO training images}. FreeSOLO extracts dense masks and removes redundancy via mask non-maximum-suppression (NMS). CutLER and CWM extracts at most 3 masks per image. The pseudo-masks are used as self-supervision signals for training downstream detectors. ~\pageref{figmention:segment}} 
  \label{fig:zero_shot_1}
\end{figure*}

\begin{figure*}
  \centering
  \includegraphics[width=0.96\linewidth]{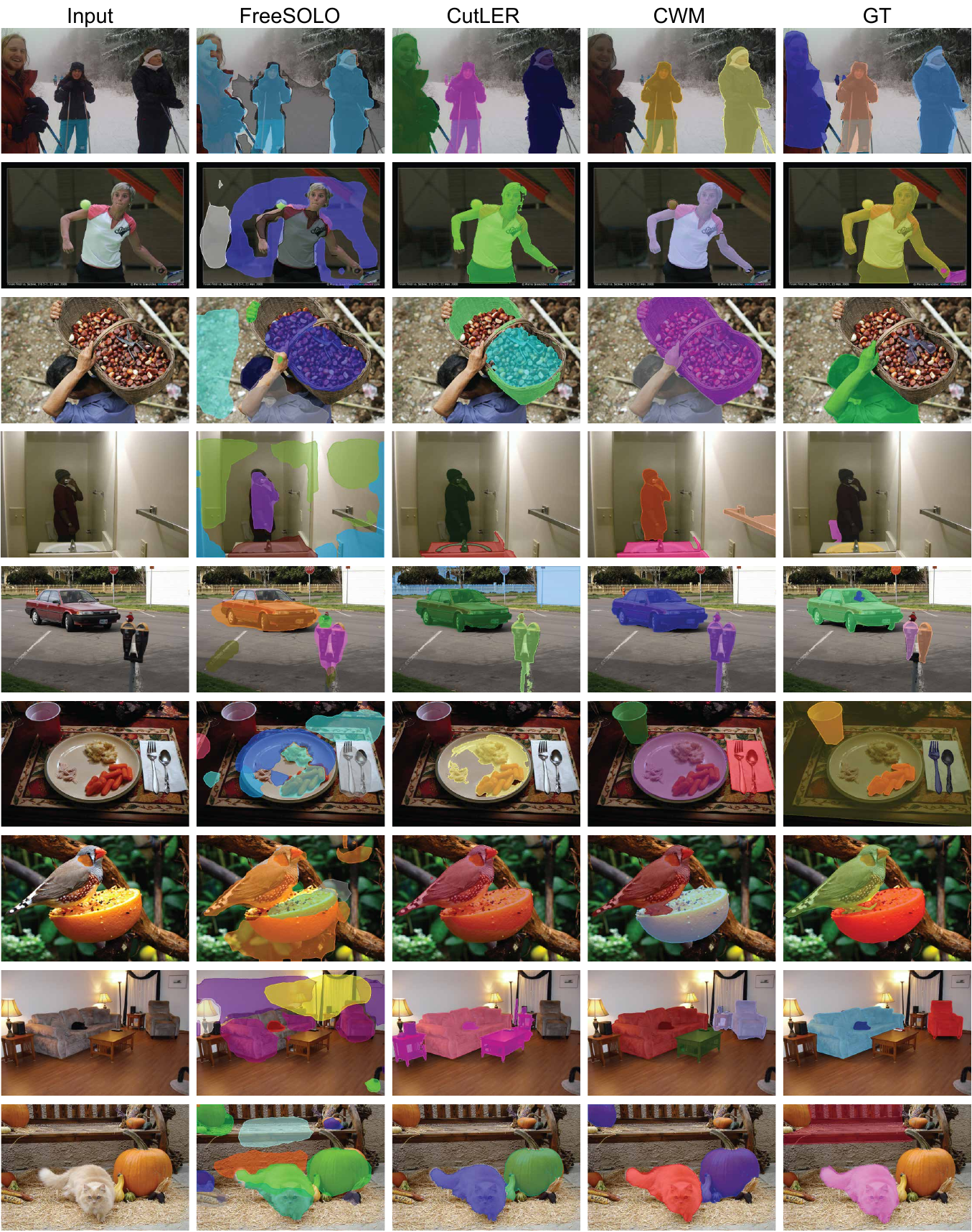}
  \caption{\label{second}\textbf{More qualitative results on pseudo-masks extracted in a zero-shot manner on COCO training images}.~\pageref{figmention:segment}} 
  \label{fig:zero_shot_2}
\end{figure*}

\end{subfigures}

\begin{subfigures}

\begin{figure*}
  \centering
  \includegraphics[width=0.98\linewidth]{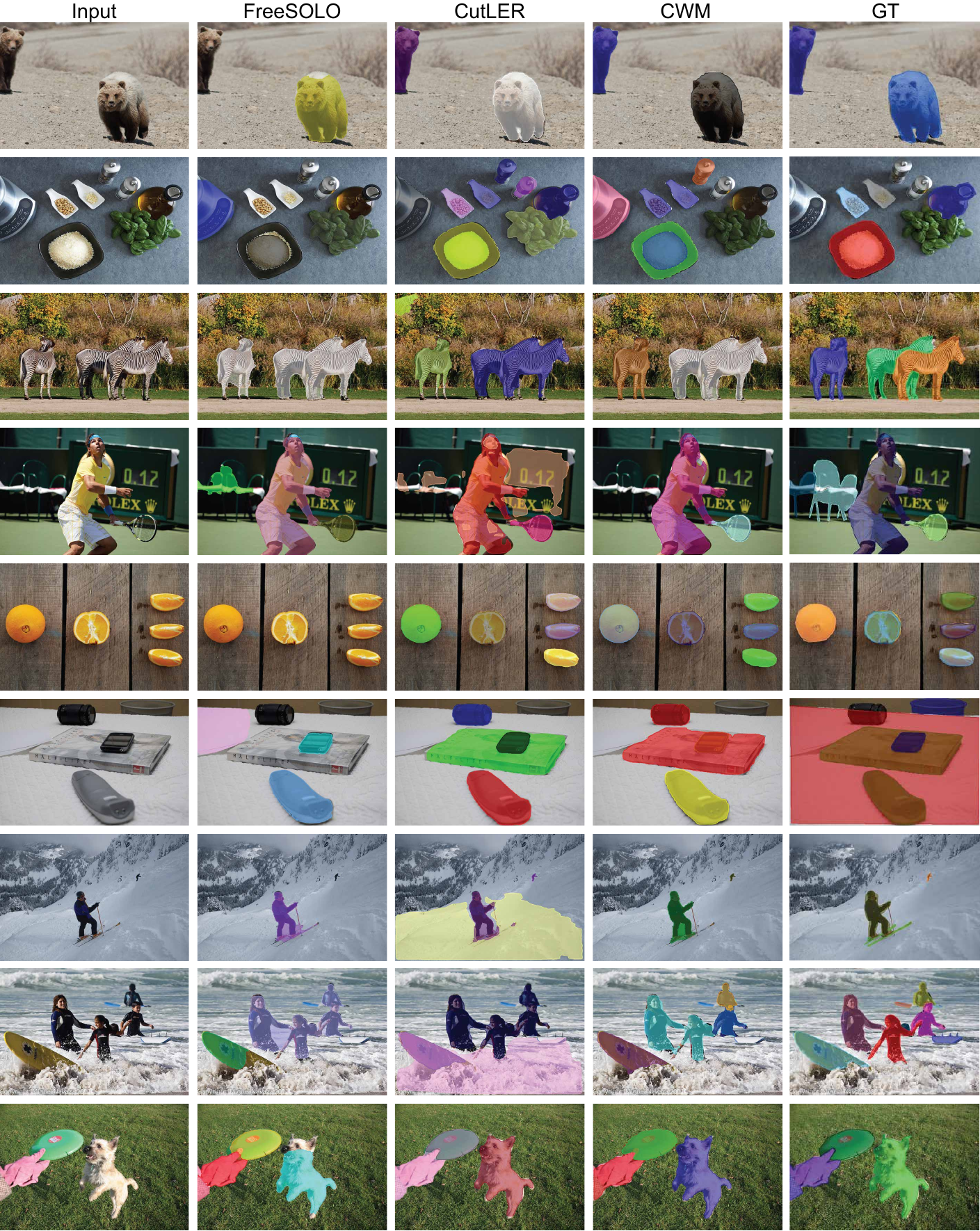}
  \caption{\label{first}\textbf{Unsupervised segmentation results from the distilled instance segmenter}. We show the results on COCO validation images.~\pageref{figmention:segment}} 
  \label{fig:maskrcnn_1}
\end{figure*}

\begin{figure*}
  \centering
  \includegraphics[width=0.98\linewidth]{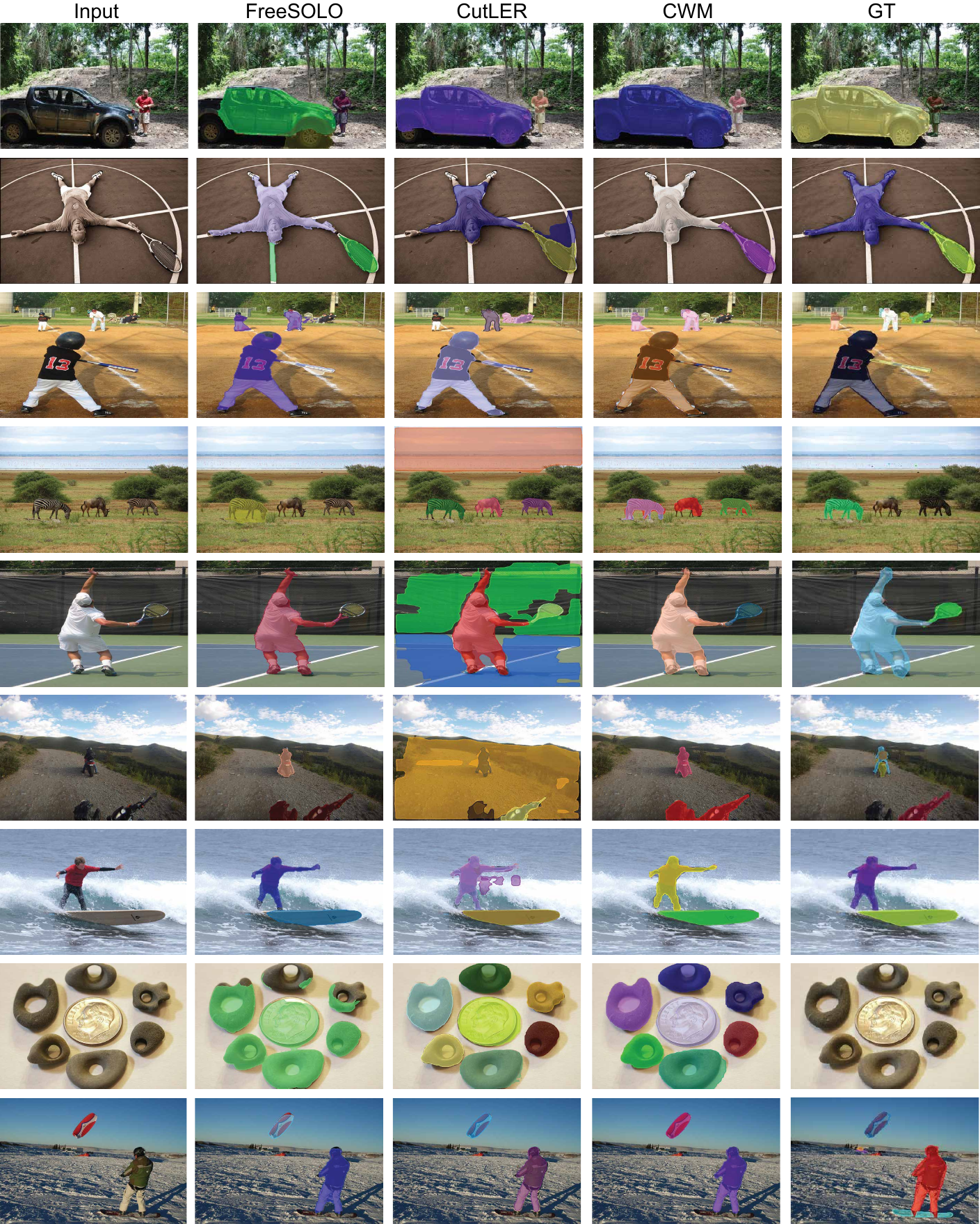}
  \caption{\label{second}\textbf{More unsupervised segmentation results from the distilled instance segmenter}.~\pageref{figmention:segment}} 
  \label{fig:maskrcnn_2}
\end{figure*}
\end{subfigures}

\clearpage
\begin{table*}[p]
  \centering
  \footnotesize
  \caption{\textbf{Physion v1.5 scenario-wise OCP results.} We compare CWM to four classes of baseline methods across different architectures on the OCP task. In the main text, we presented the scores averaged across all seven scenarios. Here we provide a detailed breakdown of performance for each specific scenario. For a strictly fair comparison we train VideoMAE* with the same patch size and number of frames as CWM.~\pageref{figmention:physion}}
  
\resizebox{\columnwidth}{!}{%

  \begin{tabular}{llC{1.2cm}C{1.2cm}C{1.2cm}C{1.2cm}C{1.2cm}C{1.2cm}C{1.2cm}c}
    \toprule
    \multirow{2}{*}{method} & \multirow{2}{*}{arch} & \multicolumn{7}{c}{OCP by scenario $\uparrow$ } & \multirow{2}{*}{Avg. OCP $\uparrow$}  \\
    \cmidrule{3-9} 
    & & collide & drop & support & link & roll & contain & dominoes &  \\
    \midrule
    \multicolumn{6}{l}{\textit{video prediction models}} \\
    \midrule
    MCVD \cite{voleti2022mcvd}  & UNet & 73.3 & 65.0 & 70.7	& 59.2 & 51.0& 66.7 & 57.9	& 63.4   \\ 
    R3M \cite{nair2022r3m}   & Res50  & 79.0 & 61.8 & 70.1 & 66.9	&\textbf{ 66.2}	& 62.1 & 67.3 & 67.6   \\
    FitVid \cite{babaeizadeh2021fitvid}  &  VAE & 65.0 & 68.2 & 71.5 & 59.9 & 54.1 & 60.8 & 70.7 & 64.3   \\
    TECO \cite{yan2023temporally}  & vq-gan  & 75.9 & 70.6 & 74.4 & 65.2 & 59.1 & 72.4 & 67.5 & 69.3   \\
    \midrule
    \multicolumn{6}{l}{\textit{self-supervised image representation models}} \\
     \midrule
    DINO \cite{caron2021emerging} & ViT-B  & 79.0 & 72.6 & 81.0 & \textbf{72.0} & 61.8 & 69.3 & 69.2 & 72.1   \\
    DINOv2 \cite{oquab2023dinov2}  & ViT-B & 78.1 & 70.7 & 80.3 & 70.7 & 64.3 & 70.6 & 70.4 & 72.2 \\
    DINOv2 \cite{oquab2023dinov2}  & ViT-L  & 81.0 & 68.8 & 82.3 & 68.8 & 61.1	& 69.9 & 73.6 & 72.2 \\
    DINOv2 \cite{oquab2023dinov2}  & ViT-g  & 80.0	& 74.5 & 81.0 & 66.2 & 61.8 &	74.5 & 71.1 & 72.7 \\
    MAE \cite{he2022masked}  & ViT-B  & 80.0 & 72.6 & 78.9 & 70.1 & 64.3 & 68.0 & 74.2 & 72.6  \\
    MAE \cite{he2022masked}  & ViT-L  & 80.0 & 73.2 & 81.0 & 69.4 & 58.0 & 69.3 & 70.4 & 71.6 \\
    MAE \cite{he2022masked}  & ViT-H & \textbf{83.8} & 72.0 & \textbf{84.4} & 70.1 & 61.8 & 69.3 & 71.7 & 73.3   \\
    MAE \cite{he2022masked} & ViT-B  & 81.9	& 70.7 & 83.0 & 68.8 & 59.9 & 67.3 & 73.0 & 72.1   \\
    MAE \cite{he2022masked} & ViT-L  & 83.8 & 70.7 & 81.0 & 68.2 & 59.9 & 70.6	& 74.2 & 72.6 \\
    
    \midrule
    
    \multicolumn{6}{l}{\textit{self-supervised video representation models}} \\
     \midrule
    VMAE \cite{tong2022videomae}  & ViT-B & 74.3 & 74.5 & 83.0	& 65.6	& 61.8	& 71.2 & 74.2 & 72.1  \\
    VMAE* \cite{tong2022videomae} & ViT-B  & 80.0 & 71.3 & 82.3 & 70.1 & 58.6 & 72.5 & 76.7 & 73.2  \\
    VMAE \cite{tong2022videomae}  & ViT-L  & 79.0 & 73.9 & 82.3 & 66.9	& 65.0 & 72.5 & 75.5 & 73.6  \\
    VMAE \cite{tong2022videomae}  & ViT-H  & 81.0 & 73.2 & 81.6 & 70.7	& 63.1 & 	70.6 & 74.2 & 73.5  \\
    VMAEv2 \cite{wang2023videomae}  & ViT-g & 77.1 & 75.2 & 83.0 & 65.0 & 62.4 & 70.6 & 72.3	& 72.2  \\
    V-JEPA \cite{tong2022videomae} & ViT-L  & 80.1 & 68.8 & 84.3 &  69.4 & 62.4 & 73.8 & 74.2 & 73.4  \\
    
    \midrule
    \multicolumn{6}{l}{\textit{vision-language models}} \\
    \midrule
    GPT4-V \cite{OpenAI2023GPT4Vision}  & - &  52.7 &  46.5 &  58.5 &  54.8 &  56.2 &  56.1 &  46.5 &  52.9  \\
    \midrule
    \midrule
    CWM  & ViT-B & 82.9 & 75.2 & 83.7 & 70.7 & 63.7 & \textbf{77.8} & 77.4 &  75.9  \\
    CWM  & ViT-L  & \textbf{83.8} & 74.5 & \textbf{84.4} & 71.3 & 65.0 & 75.8 & \textbf{78.0}	&  \textbf{76.1}  \\
    
    \bottomrule
  \end{tabular}
  }

  \label{tab:main_results_ocp_sce}
  \vspace{-0.3cm}
\end{table*}
\clearpage

\begin{table*}[p]
  \centering
  \footnotesize
  \caption{\textbf{Physion v1.5 scenario-wise OCD results.} We compare CWM to four classes of baseline methods across different architectures on the OCD task. In the main text, we presented the scores averaged across all seven scenarios. Here we provide a detailed breakdown of performance for each specific scenario. For a strictly fair comparison we train VideoMAE* with the same patch size and number of frames as CWM.~\pageref{figmention:physion}}
  
  \resizebox{\columnwidth}{!}{%
  \begin{tabular}{llC{1.2cm}C{1.2cm}C{1.2cm}C{1.2cm}C{1.2cm}C{1.2cm}C{1.2cm}c}
    \toprule
    \multirow{2}{*}{method} & \multirow{2}{*}{arch} & \multicolumn{7}{c}{OCD by scenario $\uparrow$ } & \multirow{2}{*}{Avg. OCD $\uparrow$}  \\
    \cmidrule{3-9} 
    & & collide & drop & support & link & roll & contain & dominoes &  \\
    \midrule
    \multicolumn{6}{l}{\textit{video prediction models}} \\
    \midrule
    MCVD \cite{voleti2022mcvd} & UNet   & 82.9 & 74.5 & 95.9 & 75.8 & 68.8 & 77.8 & 89.9 & 80.8 \\ 
    R3M \cite{nair2022r3m}   & Res50 & 83.8 & 72.0 & 90.5 & 72.0 & 72.0 & 70.6 & 86.2 & 78.1 \\
    FitVid \cite{babaeizadeh2021fitvid}   &  VAE  & 58.9 &  56.7 &  63.1 &  63.2 &  60.2 &  55.5 &  58.8 & 59.5 \\
    TECO \cite{yan2023temporally} & vq-gan  & 87.0 & 77.5 & 87.5 & 70.7 & 72.6 & 76.3 & 95.0 & 80.9 \\
    \midrule
    \multicolumn{6}{l}{\textit{self-supervised image representation models}} \\
     \midrule
    DINO \cite{caron2021emerging}  & ViT-B & 87.6 & 79.6 & 95.2 & 81.5 & 76.4 & 80.4 & \textbf{96.9} & 85.4 \\
    DINOv2 \cite{oquab2023dinov2}  & ViT-B  & 89.5 & \textbf{84.7} & \textbf{96.6} & \textbf{86.6} & 76.4 & 79.1 & \textbf{96.9} & 87.1 \\
    DINOv2 \cite{oquab2023dinov2}  & ViT-L  & 91.4 & 79.0 & \textbf{96.6} & 84.1 & 73.9 & 77.8 & 95.6 & 85.5 \\
    DINOv2 \cite{oquab2023dinov2} & ViT-g & 91.4 & 80.3 & 95.2 & 83.4 & 70.1 & 77.1 & 94.3 & 84.6\\
    MAE \cite{he2022masked}  & ViT-B  & 86.7 & 76.4 & 92.5 & 77.7 & 71.3 & 72.5 & 93.7 & 81.6\\
    MAE \cite{he2022masked}  & ViT-L  & 86.7 & 75.8 & 93.9 & 80.3 & 70.7 & 73.2 & 95.6 & 82.3 \\
    MAE \cite{he2022masked} & ViT-H  & 84.8 & 75.2 & 92.5 & 76.4 & 67.5 & 73.2 & 96.2 &  80.8 \\
    MAE \cite{he2022masked}  & ViT-B & 86.7 & 75.8 & 91.2 & 76.4 & 70.7 & 74.5 & \textbf{96.9} & 81.7 \\
    MAE \cite{he2022masked}  & ViT-L & 89.5 & 76.4 & 93.2 & 77.1 & 69.4 & 72.5 & 95.0 & 81.9 \\
    
    \midrule
    
    \multicolumn{6}{l}{\textit{self-supervised video representation models}} \\
     \midrule
    VMAE \cite{tong2022videomae} & ViT-B  & 91.4 & 78.3 & 94.6 & 80.3 & 76.4 & 83.0 & 95.6 & 85.7  \\
    VMAE* \cite{tong2022videomae} & ViT-B  & 93.3 & 79.6 & 94.6 & 82.2 & 75.2 & 82.4 & 96.2 & 86.2  \\
    VMAE \cite{tong2022videomae}  & ViT-L  & 95.2 & 78.3 & 95.2 & 82.8 & 75.2 & 80.4 & 95.6 &  86.1 \\
    VMAE \cite{tong2022videomae} & ViT-H & 95.2 & 79.6 & 95.2 & 84.7 & 79.0 & 81.7 & \textbf{96.9} &  87.5 \\
    VMAEv2 \cite{wang2023videomae} & ViT-g & 91.4 & 78.3 & 91.8 & 84.7 & 73.9 & 81.7 & 93.1 & 85.0 \\
    V-JEPA \cite{tong2022videomae} & ViT-L  & 93.3 & 84.7 & 95.9 & 83.4 & 73.2 & 83.0 & 95.6 & 87.0  \\
    \midrule
    \multicolumn{6}{l}{\textit{vision-language models}} \\
    \midrule
    GPT4-V \cite{OpenAI2023GPT4Vision}  & - & 52.9 & 53.0 & 54.2 & 60.7 & 56.2 & 56.1 & 49.7 & 54.7 \\
    \midrule
    \midrule
    CWM  & ViT-B & 96.2 & 82.2 & 95.9 & 85.4 & 81.5 & \textbf{86.3 }& 96.2 & \textbf{89.1} \\
    CWM & ViT-L & \textbf{96.2} & 83.4 & \textbf{96.6} & 84.1 & 81.5 & 83.0 & 96.2 & 88.7\\
    
    \bottomrule
  \end{tabular}
  }
 
  \label{tab:main_results_ocd_sce}
  \vspace{-0.2cm}
\end{table*}

\end{document}